%% file: main.tex
\pdfoutput=1

\documentclass[11pt]{article}

\usepackage{acl}

\usepackage{times}
\usepackage{latexsym}

\usepackage[T1]{fontenc}

\usepackage[utf8]{inputenc}

\usepackage{microtype}

\title{\mytool{}: Benchmark, Reevaluation, Reflections, and \\ Future Challenges in Event Extraction}

\author{Kuan-Hao Huang$^{\dagger}$ \ \ \ 
I-Hung Hsu$^{\diamond}$ \ \ \ Tanmay Parekh$^{\ddagger}$ \ \ \ 
Zhiyu Xie$^{\oplus}$ 
\ \ \ Zixuan Zhang$^{\dagger}$ \\
{\bf Premkumar Natarajan$^{\diamond}$ \ \ \ 
Kai-Wei Chang$^{\ddagger}$ \ \ \ 
Nanyun Peng$^{\ddagger}$ \ \ \ 
Heng Ji$^{\dagger}$} \vspace{0.3em}\\
$^{\dagger}$University of Illinois Urbana-Champaign \ \ \ 
$^{\diamond}$University of Southern California \\ 
$^{\ddagger}$University of California, Los Angeles \ \ \ $^{\oplus}$Stanford University \vspace{0.3em} \\
\texttt{\{khhuang, zixuan11, hengji\}@illinois.edu} \\
\texttt{\{ihunghsu, premkumn\}@usc.edu} \ \ \
\texttt{zhiyuxie@stanford.edu}\\
\texttt{\{tparekh, kwchang, violetpeng\}@cs.ucla.edu} \\
}

\input{macros}

\begin{document}
\maketitle
\input{00-abstract}

\input{01-introduction}

\input{02-related.tex}
\input{03-past}

\input{04-textee}

\input{05-llm}

\input{06-future}

\input{10-conclusion}

\input{11-limitation}

\bibliography{custom}

\clearpage
\appendix
\input{90-datasets}
\input{91-models}
\input{92-results}

\input{93-prompt}

\end{document}

%% file: macros.tex
\usepackage{graphicx}
\usepackage{multirow}
\usepackage{makecell}
\usepackage{booktabs}
\usepackage{enumitem}
\usepackage{amssymb}
\usepackage{xcolor}
\usepackage{colortbl}
\usepackage{subcaption}
\usepackage{threeparttable}
\definecolor{dark-green}{rgb}{0.31, 0.47, 0.26}
\definecolor{dark-red}{rgb}{0.81, 0.09, 0.13}
\usepackage{amsmath}
\usepackage{soul}
\setuldepth{Berlin}
\usepackage{tabularx}
\newcolumntype{x}[1]{>{\centering\arraybackslash\hspace{0pt}}m{#1}}
\newcolumntype{y}[1]{>{\arraybackslash\hspace{0pt}}m{#1}}

\usepackage{pifont}
\newcommand{\cmark}{\color{dark-red}{\ding{51}}}

\newcommand{\mypar}[1]{\vspace{0.25em}\noindent\textbf{#1}}

\newcommand{\mytool}{\textsc{TextEE}}

\usepackage{todonotes}

%% file: 00-abstract.tex
\begin{abstract}
Event extraction has gained considerable interest due to its wide-ranging applications. 
However, recent studies draw attention to evaluation issues, suggesting that reported scores may not accurately reflect the true performance.
In this work, we identify and address evaluation challenges, including \emph{inconsistency} due to varying data assumptions or preprocessing steps, the \emph{insufficiency} of current evaluation frameworks that may introduce dataset or data split bias, and the \emph{low reproducibility} of some previous approaches.
To address these challenges, we present \mytool{}, a standardized, fair, and reproducible benchmark for event extraction.
\mytool{} comprises standardized data preprocessing scripts and splits for 16 datasets spanning eight diverse domains
and includes 14 recent methodologies, conducting a comprehensive benchmark reevaluation.
We also evaluate five varied large language models on our \mytool{} benchmark and demonstrate how they struggle to achieve satisfactory performance.
Inspired by our reevaluation results and findings, we discuss the role of event extraction in the current NLP era, as well as future challenges and insights derived from \mytool{}. 
We believe \mytool{}, the first standardized comprehensive benchmarking tool, will significantly facilitate future event extraction research.\footnote{\mytool{}  benchmark platform is available at \url{https://github.com/ej0cl6/TextEE}}
\end{abstract}

%% file: 01-introduction.tex
\section{Introduction}
\label{sec:intro}

Event extraction~\cite{ji2008refining} has always been a challenging task in the field of natural language processing (NLP) due to its demand for a high-level comprehension of texts.
Since event extraction benefits many applications \cite{Zhang20app1,Han21app2}, it has attracted increasing attention in recent years \cite{Luan19dygie,Lin20oneie,Nguyen21fourie,Hsu22degree,Ma22paie}.
However, due to the complicated nature of event extraction datasets and systems, fairly evaluating and comparing different event extraction approaches is not straightforward.
Recent attempts \cite{Zheng21reeval,Peng23omnievent,Peng23pitfall} point out that the reported scores in previous work might not reflect the true performance in real-world applications because of various shortcomings and issues during the evaluation process.
This poses a potential obstacle to the development of robust techniques for research in event extraction.

Motivated by the evaluation concern, this work aims to establish a standardized, fair, and reproducible benchmark for assessing event extraction approaches.
We start by identifying and discussing several significant issues in the current evaluation process.
First, we discuss the \emph{inconsistency} issue caused by discrepant assumptions about data, different preprocessing steps, and the use of external resources. 
Next, we highlight the \emph{insufficiency} problem of existing evaluation pipelines,  which cover limited datasets and rely on fixed data splits, potentially introducing bias when evaluating performance.
Finally, we emphasize the importance of \emph{reproducibility}, which indirectly causes the aforementioned inconsistency and insufficiency issues.

To address these evaluation concerns, we propose \mytool{}, an evaluation platform that covers 16 datasets spanning diverse domains.
To ensure fairness in comparisons, we standardize data preprocessing procedures and introduce five standardized data splits.
Furthermore, we aggregate and re-implement 14 event extraction approaches published in recent years and conduct a comprehensive reevaluation.
\mytool{} offers the benefits of \emph{consistency}, \emph{sufficiency}, \emph{reproducibility} in evaluation.
Additionally, we benchmark several large language models (LLMs) \cite{touvron2023llama,Tunstall23zephyr,Jiang24mixtral} for event extraction with \mytool{} and show the unsatisfactory performance of LLMs for this task. 

Based on our reevaluation results and findings, we discuss the role of event extraction in the current era of LLMs, along with challenges and insights gleaned from \mytool{}. 
Specifically, we discuss how event extraction systems can be optional tools for LLMs to utilize, as well as highlight future challenges, including enhancing generalization, expanding event coverage, and improving efficiency.

In summary, our contributions are as follows:
(1) We highlight and address the difficulties of fair evaluation for event extraction tasks.
(2) We present \mytool{} as a benchmark platform for event extraction research and conduct a thorough reevaluation of recent approaches as well as LLMs.
(3) Based on our results and findings, we discuss limitations and future challenges in event extraction.

%% file: 02-related.tex
\section{Background and Related Work}

\subsection{Event Extraction}

\begin{figure}[t]
    \centering
    \includegraphics[width=0.48\textwidth]{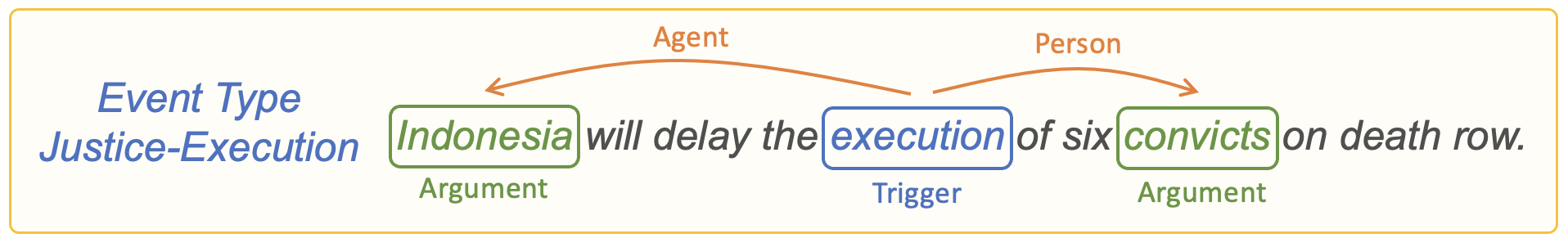}
    \caption{An example of a \emph{Justice-Execution} event. One trigger span (\emph{execution}) and two argument roles, \emph{Indonesia (Agent)} and \emph{convicts (Person)}, are identified.}
    \label{fig:example}
    \vspace{-1em}
\end{figure}

Event extraction (EE) aims to identify structured information from texts. 
Each event consists of an event type, a trigger span, and several arguments along with  their roles.\footnote{In this work, we only cover closed-domain EE with a given ontology. We consider event mentions as events and do not consider event coreference resolution.}
Figure~\ref{fig:example} shows an example of a \emph{Justice-Execution} event extracted from the text.
This event is triggered by the text span \emph{execution} and contains two argument roles, including \emph{Indonesia (Agent)} and \emph{convicts (Person)}.

Previous work can be categorized into two types:
(1) \textbf{End-to-end (E2E)} approaches extract event types, triggers, and argument roles in an end-to-end manner.
(2) Pipeline approaches first solve the \textbf{event detection (ED)} task, which detects trigger spans and the corresponding event types, then deal with the \textbf{event argument extraction (EAE)} task, which extracts arguments and the corresponding roles given an event type and a trigger span.

\subsection{Related Work}
\label{sec:related-work}

\mypar{Event extraction.}
Most end-to-end approaches construct graphs to model the relations between entities and extract triggers and argument roles accordingly \cite{Luan19dygie,Wadden19dygiepp,Han19add3,Lin20oneie,Huang20add2,Nguyen21fourie,Zhang21amrie,Huang21add1}. 
There is a recent focus on employing generative models to generate summaries for extracting events \cite{text2event,Hsu22degree}.
Unlike end-to-end approaches, pipeline methods train two separate models for event detection and event argument extraction.
Different techniques are introduced, such as question answering \cite{Du20bertqa,Liu20rcee,Li20multi,Lu23edqa}, language generation \cite{Paolini21tanl,Hsu22degree}, querying and extracting \cite{Wang22queryextract}, pre-training \cite{Wang20pretrain}, and multi-tasking \cite{Lu22multi,Wang23multi2}.
Some works focus on zero-shot or few-shot settings \cite{Huang18zero1,Hsu22degree}.

\mypar{Event detection.}
There are many prior studies focusing on extracting triggers only.
Most works pay attention to the standard supervised setting \cite{Liu18ed1,Lai20ed2,Veyseh21ed3,Li21ed4,Huang22unist,Liu22saliency,Li23glen}.
Some others study the few-shot setting \cite{Deng20fewed4,Zhao22fewed2,Zhang22fewed3,Ma23fewed1,Wang23apex}

\mypar{Event argument extraction.}
Event argument extraction has caught much attention in recent years \cite{Veyseh22opt,Li21wikievents,Hsu22tagprime,Zeng22eae1,Ma22paie,Huang22xgear,Xu22eae4,Hsu23ampere,Nguyen23speae,He23eae2,Huang23eae3,Parekh2023contextual}.
Some works focus on training models with only a few examples \cite{Sainz22feweae2,Yang23feweae1,Wang23code4struct}.

\mypar{Event extraction datasets.}
Most of event extraction datasets come from Wikipedia and the news domain \cite{muc1992muc4,Doddington04ace,Song15ere,Ebner20rams,Li20m2e2,Li21wikievents,Veyseh22mee,Li22clipevent}.
To increase the event type coverage, some works focus on general domain datasets \cite{Wang20maven,Deng20fewevent,parekh2023geneva,Li23glen}.
Recently, datasets in specific domains have been proposed, including cybersecurity \cite{Satyapanich20casie,Trong20cyber}, pharmacovigilance \cite{Sun22phee}, epidemic \cite{Parekh2024speed}, and historical text \cite{Lai21his}.

\mypar{Event extraction evaluation and analysis.}
Recently, some works point out several pitfalls when training event extraction models and attempt to provide solutions \cite{Zheng21reeval,Peng23omnievent,Peng23pitfall}.
Our observation partially echos their findings, while our proposed \mytool{} covers more diverse datasets and includes more recent approaches.
On the other hand, some studies discuss ChatGPT's performance on event extraction but only for one dataset \cite{Li23gpteval,Gao23chatgptee}.

%% file: 03-past.tex
\section{Issues in Past Evaluation}
\label{sec:past}

Despite a wide range of works in EE, we identify several major issues of the past evaluation.
We classify those issues into three categories: \emph{inconsistency}, \emph{insufficiency}, and \emph{low reproducibility}.

\mypar{Inconsistency.}
Due to the lack of a standardized evaluation framework, we notice that many studies utilize varied experimental setups while comparing their results with reported numbers in the literature. 
This leads to unfair comparisons and makes the evaluation less reliable and persuasive.
We identify and summarize the underlying reasons as follows:

\begin{itemize}[topsep=0pt, itemsep=-4pt, leftmargin=12pt]
    \item \textbf{Different assumptions about data.}
    In the past, different approaches tend to have their own assumptions about data.
    For instance, some works allow trigger spans consisting of multiple words \cite{Lin20oneie,Hsu22degree,Hsu22tagprime}, whereas others consider only single-word triggers \cite{Liu20rcee,Du20bertqa,Wang22queryextract};
    some studies assume that there are no overlapping argument spans \cite{Zhang21amrie}, while others can handle overlapping spans \cite{Wadden19dygiepp,Huang22xgear};
    some methods filter out testing data when the texts are too long \cite{Liu22saliency}, while others do not \cite{Hsu23ampere,Ma22paie}.
    Due to these discrepancies in assumptions, the reported numbers from the original papers are actually not directly comparable.

    \item \textbf{Different data preprocessing steps.}
    Many previous works benchmark on the ACE05 \cite{Doddington04ace} and RichERE \cite{Song15ere} datasets.
    Since these datasets are behind a paywall and not publicly accessible, people can only share the data preprocessing scripts.
    Unfortunately, we observe that some popular preprocessing scripts can generate very different data.
    For instance, the processed ACE05 datasets from \citet{Wadden19dygiepp}, \citet{Li20multi}, and \citet{Veyseh22opt} have varying numbers of role types (22, 36, and 35 respectively).
    In addition, it is crucial to note that variations in Python package versions can lead to different generated data even when using the same script.
    For example, different versions of \texttt{nltk} packages may have discrepancies in sentence tokenization and word tokenization, resulting in different processed data. 
    Such differences in preprocessing largely affect model evaluation, leading to significant discrepancies (e.g., over 4 F1 score), thereby reducing persuasiveness \cite{Peng23pitfall}.
    
    \item \textbf{Different external resources.}
    We notice that many approaches utilize additional resources without clearly describing the differences in experimental settings.
    For example, \citet{Wang23apex} employs part-of-speech tags for event detection; \citet{Sainz22nli} and \citet{Wang22queryextract} consider gold entity annotations for event argument extraction.
    These setting differences can lead to potentially unfair comparisons.
\end{itemize}


\begin{table*}[t!]
\centering
\setlength{\tabcolsep}{3pt}
\resizebox{1.0\textwidth}{!}{
\begin{tabular}{l|c|cccccc|ccc|c}
    \toprule
    Dataset & Task & \#Docs & \#Inst & \#ET & \#Evt & \#RT & \#Arg & Event & Entity & Relation & Domain \\
    \midrule
    ACE05 \cite{Doddington04ace}     & E2E, ED, EAE &   599 & 20920 &  33 &  5348 &  22 &  8097 & \cmark & \cmark & \cmark & News \\
    RichERE \cite{Song15ere}         & E2E, ED, EAE &   288 & 11241 &  38 &  5709 &  21 &  8254 & \cmark & \cmark & \cmark & News \\
    MLEE \cite{Pyysalo12mlee}        & E2E, ED, EAE &   262 &   286 &  29 &  6575 &  14 &  5958 & \cmark & \cmark & \cmark & Biomedical \\
    Genia2011 \cite{Kim11genia2011}  & E2E, ED, EAE &   960 &  1375 &   9 & 13537 &  10 & 11865 & \cmark & \cmark & & 
    Biomedical \\
    Genia2013 \cite{Kim13genia2013}  & E2E, ED, EAE &    20 &   664 &  13 &  6001 &   7 &  5660 & \cmark & \cmark & \cmark & Biomedical \\
    M$^2$E$^2$ \cite{Li20m2e2}       & E2E, ED, EAE &  6013 &  6013 &   8 &  1105 &  15 &  1659 & \cmark & \cmark & & Multimedia \\
    CASIE \cite{Satyapanich20casie}  & E2E, ED, EAE &   999 &  1483 &   5 &  8469 &  26 & 22575 & \cmark & & & Cybersecurity \\
    PHEE \cite{Sun22phee}            & E2E, ED, EAE &  4827 &  4827 &   2 &  5019 &  16 & 25760 & \cmark & & & Pharmacovigilance \\
    MAVEN \cite{Wang20maven}         & ED           &  3623 & 40473 & 168 & 96897 &  -- &  --   & \cmark & & & General \\
    FewEvent \cite{Deng20fewevent}   & ED           & 12573 & 12573 & 100 & 12573 &  -- &  --  & \cmark & & & General \\
    SPEED \cite{Parekh2024speed}     & ED           &  1975 &  1975 &   7 &  2217 &  -- &  --   & \cmark & & & Epidemic \\
    MEE \cite{Veyseh22mee}           & ED           & 13000 & 13000 &  16 & 17257 &  -- &  --   & \cmark & \cmark & & Wikipedia \\
    WikiEvents \cite{Li21wikievents} & EAE          &   245 &   565 &  50 &  3932 &  58 &  5501 & \cmark & \cmark & & Wikipedia \\
    RAMS \cite{Ebner20rams}          & EAE          &  9647 &  9647 & 139 &  9647 &  65 & 21206 & \cmark & \cmark & & News \\
    MUC-4 \cite{muc1992muc4}         & EAE          &  1700 &  2360 &   1 &  2360 &   5 &  4776 & \cmark & & & News \\
    GENEVA \cite{parekh2023geneva}   & EAE          &   262 &  3684 & 115 &  7505 & 220 & 12314 & \cmark & \cmark & & General \\
    \bottomrule

\end{tabular}}
\caption{\mytool{} supports fourteen datasets across various domains. \emph{\#Docs, \#Inst, \#ET, \#EvT, \#RT, and \#Arg represent the number of documents, instances, event types, events, roles, and arguments, respectively. Event, Entity, and Relation indicate if the dataset contains the corresponding annotations.}}
\label{tab:dataset}
\vspace{-1em}
\end{table*}

\mypar{Insufficiency.}
We argue that the existing evaluation process used by the majority of approaches cannot thoroughly evaluate the capabilities of event extraction models due to the following aspects:

\begin{itemize}[topsep=0pt, itemsep=-4pt, leftmargin=12pt]
    \item \textbf{Limited dataset coverage.}
    Early works usually utilize ACE05 \cite{Doddington04ace} and RichERE \cite{Song15ere} as the evaluation datasets.
    Consequently, most follow-up works adopt the same two datasets for comparison regardless that several new datasets across diverse domains are proposed \cite{Li21wikievents,Sun22phee,Tong22docee,parekh2023geneva}. 
    The limited dataset coverage may introduce domain bias and lead to biased evaluations.
    
    \item \textbf{Data split bias.}
    Although many works address model randomness by averaging multiple experimental runs \cite{Zhang21amrie,Hsu22degree,Wang22queryextract}, they often overlook randomness in data splits and report numbers only for a \textit{single} and \textit{fixed} split for train, dev, and test sets. This can lead to a notable bias, especially for event extraction where there is a high variance of annotation density across sentences or documents.
    For example, following the preprocessing step of \citet{Wadden19dygiepp} applied to ACE05, the resulting processed dataset has 33 event types in the train set, 21 event types in the dev set, and 31 event types in the test set. 
    Accordingly, it is likely to have a significant performance discrepancy between the dev and the test set, making the reported numbers biased.
    
\end{itemize}

\mypar{Low reproducibility.}
Because of the complex nature of event extraction tasks, the event extraction models have become increasingly complicated.
Releasing code and checkpoints for reproducing results has become essential, as many details and tricks need to be taken into account during the re-implementation process.
However, many promising approaches do not provide an official codebase \cite{Li20multi,Nguyen21fourie,Wei21feae,LiuH22gtee}, which potentially impedes the progress of research in event extraction.

%% file: 04-textee.tex
\section{Benchmark and Reevaluation}
\label{sec:textee}


\begin{table*}[t!]
\centering
\setlength{\tabcolsep}{5pt}
\resizebox{.93\textwidth}{!}{
\begin{tabular}{l|c|ccccccc}
    \toprule
    Model & Task & Event & Entity & Relation & POS Tags & AMR & Verbalization & Template \\
    \midrule
    \multicolumn{6}{l}{\emph{Classification-Based Models}} \\
    \midrule
    DyGIE++ \cite{Wadden19dygiepp} & E2E & \cmark & \cmark & \cmark \\
    OneIE \cite{Lin20oneie} & E2E & \cmark & \cmark & \cmark \\
    AMR-IE \cite{Zhang21amrie} & E2E & \cmark & \cmark & \cmark & & \cmark \\
    EEQA \cite{Du20bertqa} & ED, EAE & \cmark & & & & & & \cmark \\
    RCEE \cite{Liu20rcee} & ED, EAE & \cmark & & & & & & \cmark \\
    Query\&Extract \cite{Wang22queryextract} & ED, EAE & \cmark & & & \cmark &  & \cmark & \\
    TagPrime-C \cite{Hsu22tagprime} & ED, EAE & \cmark & & &&  & \cmark & \\
    TagPrime-CR \cite{Hsu22tagprime} & EAE & \cmark & & & & & \cmark & \\
    UniST \cite{Huang22unist} & ED & \cmark & & &&  & \cmark & \\
    CEDAR \cite{Li23glen} & ED & \cmark & & &&  & \cmark & \\
    \midrule
    \multicolumn{6}{l}{\emph{Generation-Based Models}} \\
    \midrule
    DEGREE \cite{Hsu22degree} & E2E, ED, EAE & \cmark & & & & & \cmark & \cmark \\
    BART-Gen \cite{Li21wikievents} & EAE & \cmark & & & & & & \cmark \\
    X-Gear \cite{Huang22xgear} & EAE & \cmark \\
    PAIE \cite{Ma22paie} & EAE & \cmark & & & & & \cmark & \cmark \\
    AMPERE \cite{Hsu23ampere} & EAE & \cmark & & & & \cmark & \cmark & \cmark \\
    \bottomrule

\end{tabular}}
\caption{\mytool{} supports various models with different assumptions. \emph{Event, Entity, Relation, POS Tags, and AMR indicate if the model considers the corresponding annotations. Verbalization: if the model requires verbalized type strings. Template: if the model needs a human-written template to connect the semantics of triggers and arguments.}}
\label{tab:models}
\vspace{-1em}
\end{table*}

To address the issues listed in Section~\ref{sec:past}, we present \mytool{}, a framework aiming to standardize and benchmark the evaluation process of event extraction.
\mytool{} has several advantages as follows.

\mypar{Better Consistency.}
We propose a standardized experimental setup for fair comparisons.

\begin{itemize}[topsep=0pt, itemsep=-4pt, leftmargin=12pt]
    \item \textbf{Normalizing assumptions about data.}
    We adopt the loosest assumption about data to align with real-world cases effectively.
    This includes allowing multiple-word triggers, considering overlapping argument spans, and retaining all instances without filtering.

    \item \textbf{Standardizing data preprocessing steps.}
    We provide a standard script for data preprocessing, including tokenization and label offset mapping.
    To avoid the difference caused by variations in Python package versions, we use \texttt{stanza 1.5.0} for tokenization and save all the offsets.
    Our script will load the saved offsets during preprocessing, ensuring that everyone can generate exactly the same data.
    
    \item \textbf{Specifying additional resources.}
    We clearly specify the resources utilized by all baselines (Table~\ref{tab:models}).
    For approaches that require additional gold annotations (such as POS tags, AMR, and gold entities), considering the purpose of fair comparisons, we either train a new predictor from training annotations (for entities) or use a pre-trained model (for POS tags and AMR), and consider the predicted labels as a substitute for the gold annotations.
\end{itemize}

\mypar{Improved Sufficiency.}
We improve the sufficiency of the evaluation process as follows.

\begin{itemize}[topsep=0pt, itemsep=-4pt, leftmargin=12pt]
    \item \textbf{Increasing dataset coverage.}
    As listed in Table~\ref{tab:dataset}, we increase the dataset coverage by including \emph{sixteen} event extraction datasets that cover various domains.
    
    \item \textbf{Providing standard data splits.}
    For each dataset, we merge all the labeled data and re-generate data splits.
    To mitigate the data split bias, we offer \emph{five} split for each dataset and report the average results.
    To reduce the distribution gap among the train, dev, and test sets, we select splits that these sets share the most similar statistics, such as the number of event types and role types, as well as the number of events and arguments.
    Appendix~\ref{app:data} lists the detailed statistics of each split for each dataset.

    \item \textbf{New evaluation metrics.}    
    Most prior works follow \citet{Lin20oneie} and consider Trigger F1-score and Argument F1-score as the evaluation metrics.
    Specifically, they calculate F1-scores regarding the following:
    (1) \textbf{TI}: if the \emph{(start\_idx, end\_idx)} of a predicted trigger match the gold ones. 
    (2) \textbf{TC}: if the \emph{(start\_idx, end\_idx, event\_type)} of a predicted trigger match the gold ones.
    (3) \textbf{AI}: if the \emph{(start\_idx, end\_idx, event\_type)} of a predicted argument match the gold ones. 
    (4) \textbf{AC}: if the \emph{(start\_idx, end\_idx, event\_type, role\_type)} of a predicted argument match the gold ones.
    However, we notice that AI and AC cannot precisely evaluate the quality of predicted arguments.
    There can be multiple triggers sharing the same event type in an instance, but the current score \emph{does not} evaluate if the predicted argument attaches to the correct trigger.
    Accordingly, we propose two new scores to evaluate this attachment:
    (5) \textbf{AI+}: if the \emph{(start\_idx, end\_idx, event\_type, attached\_trigger\_offsets)} of a predicted argument match the gold ones. 
    (6) \textbf{AC+}: if the \emph{(start\_idx, end\_idx, event\_type, attached\_trigger\_offsets, role\_type)} of a predicted argument match the gold ones.
    
\end{itemize}

\mypar{Reproducibility.}
We open-source the proposed \mytool{} framework for better reproducibility. 
Additionally, we encourage the community to contribute their datasets and codebases to advance the research in event extraction.

\subsection{\mytool{} Benchmark}

\mytool{} supports 16 datasets across various domains and 14 models proposed in recent years.

\mypar{Dataset.}
In addition to the two most common datasets, \textbf{ACE05} \cite{Doddington04ace} and \textbf{RichERE} \cite{Song15ere}, which particularly focus on the news domain, we consider as many other event extraction datasets across diverse domains as possible, including \textbf{MLEE} \cite{Pyysalo12mlee}, \textbf{Genia2011} \cite{Kim11genia2011}, and \textbf{Genia2013} \cite{Kim13genia2013} from the biomedical domain, \textbf{CASIE} \cite{Satyapanich20casie} from the cybersecurity domain, \textbf{PHEE} \cite{Sun22phee} from the pharmacovigilance domain, \textbf{SPEED} \cite{Parekh2024speed} from the epidemic domain, \textbf{M$^2$E$^2$} \cite{Li20m2e2}, \textbf{MUC-4} \cite{muc1992muc4}, and \textbf{RAMS} \cite{Ebner20rams} from the news domain, \textbf{MEE} \cite{Veyseh22mee} and \textbf{WikiEvents} \cite{Li21wikievents} from Wikipedia, \textbf{MAVEN} \cite{Wang20maven}, \textbf{FewEvent} \cite{Deng20fewevent}, and \textbf{GENEVA} \cite{parekh2023geneva} from the general domain.
We also notice that there are other valuable datasets, such as GLEN \cite{Li23glen} and VOANews \cite{Li22clipevent}, but we do not include them as their training examples are not all annotated by humans.
Table~\ref{tab:dataset} summarizes the statistics for each dataset after our preprocessing steps. Appendix~\ref{app:data} describes the details of the preprocessing steps and our assumptions.

\begin{table*}[t!]
\centering
\setlength{\tabcolsep}{6pt}
\resizebox{.99\textwidth}{!}{
\begin{tabular}{l|cccc|cccc|cccc|cccc}
    \toprule
    \multirow{2}{*}{Model} & \multicolumn{4}{c|}{\textbf{ACE05}} & \multicolumn{4}{c|}{\textbf{RichERE}} & \multicolumn{4}{c|}{\textbf{MLEE}} & \multicolumn{4}{c}{\textbf{Genia2011}} \\
    & TI & TC & AC & AC+ & TI & TC & AC & AC+ & TI & TC & AC & AC+ & TI & TC & AC & AC+ \\
    \midrule
        DyGIE++
    & \cellcolor{red!16}74.7 & \cellcolor{red!24}71.3 & \cellcolor{red!00}56.0 & \cellcolor{red!08}51.8 
    & \cellcolor{red!08}69.7 & \cellcolor{red!00}59.8 & \cellcolor{red!00}42.0 & \cellcolor{red!00}38.3 
    & \cellcolor{red!16}82.6 & \cellcolor{red!08}78.2 & \cellcolor{red!16}57.8 & \cellcolor{red!16}54.4 
    & \cellcolor{red!00}74.2 & \cellcolor{red!00}70.3 & \cellcolor{red!08}56.9 & \cellcolor{red!16}52.1 
    \\
    OneIE
    & \cellcolor{red!24}75.0 & \cellcolor{red!08}71.1 & \cellcolor{red!16}59.9 & \cellcolor{red!24}54.7 
    & \cellcolor{red!24}71.0 & \cellcolor{red!16}62.5 & \cellcolor{red!16}50.0 & \cellcolor{red!16}45.2 
    & \cellcolor{red!24}82.7 & \cellcolor{red!16}78.5 & \cellcolor{red!00}26.9 & \cellcolor{red!00}13.1 
    & \cellcolor{red!16}76.1 & \cellcolor{red!08}72.1 & \cellcolor{red!16}57.0 & \cellcolor{red!00}33.6 
    \\
    AMR-IE
    & \cellcolor{red!08}74.6 & \cellcolor{red!16}71.1 & \cellcolor{red!24}60.6 & \cellcolor{red!08}54.6 
    & \cellcolor{red!16}70.5 & \cellcolor{red!08}62.3 & \cellcolor{red!08}49.5 & \cellcolor{red!08}44.7 
    & \cellcolor{red!08}82.4 & \cellcolor{red!08}78.2 & \cellcolor{red!00}15.2 & \cellcolor{red!00}4.7 
    & \cellcolor{red!24}76.4 & \cellcolor{red!24}72.4 & \cellcolor{red!00}42.8 & \cellcolor{red!00}29.0 
    \\
    EEQA
    & \cellcolor{red!00}73.8 & \cellcolor{red!00}70.0 & \cellcolor{red!00}55.3 & \cellcolor{red!00}50.4 
    & \cellcolor{red!00}69.3 & \cellcolor{red!00}60.2 & \cellcolor{red!00}45.8 & \cellcolor{red!00}41.9 
    & \cellcolor{red!00}81.4 & \cellcolor{red!00}76.9 & \cellcolor{red!08}51.1 & \cellcolor{red!08}38.1 
    & \cellcolor{red!08}74.4 & \cellcolor{red!08}71.3 & \cellcolor{red!08}50.6 & \cellcolor{red!08}38.4 
    \\
    RCEE
    & \cellcolor{red!08}74.0 & \cellcolor{red!08}70.5 & \cellcolor{red!00}55.5 & \cellcolor{red!00}51.0 
    & \cellcolor{red!00}68.6 & \cellcolor{red!00}60.0 & \cellcolor{red!00}46.2 & \cellcolor{red!00}42.1 
    & \cellcolor{red!00}81.3 & \cellcolor{red!00}77.2 & \cellcolor{red!00}49.3 & \cellcolor{red!00}35.4 
    & \cellcolor{red!00}73.3 & \cellcolor{red!00}70.1 & \cellcolor{red!00}49.0 & \cellcolor{red!00}37.2 
    \\
    Query\&Extract
    & \cellcolor{red!00}68.6 & \cellcolor{red!00}65.1 & \cellcolor{red!00}55.0 & \cellcolor{red!00}49.0 
    & \cellcolor{red!00}67.5 & \cellcolor{red!00}59.8 & \cellcolor{red!00}48.9 & \cellcolor{red!00}44.5 
    & \cellcolor{red!00}-- & \cellcolor{red!00}-- & \cellcolor{red!00}-- & \cellcolor{red!00}-- 
    & \cellcolor{red!00}-- & \cellcolor{red!00}-- & \cellcolor{red!00}-- & \cellcolor{red!00}-- 
    \\
    TagPrime
    & \cellcolor{red!00}73.2 & \cellcolor{red!00}69.9 & \cellcolor{red!08}59.8 & \cellcolor{red!16}54.6 
    & \cellcolor{red!08}69.6 & \cellcolor{red!24}63.5 & \cellcolor{red!24}52.8 & \cellcolor{red!24}48.4 
    & \cellcolor{red!08}81.8 & \cellcolor{red!24}79.0 & \cellcolor{red!24}65.2 & \cellcolor{red!24}60.3 
    & \cellcolor{red!08}74.9 & \cellcolor{red!16}72.2 & \cellcolor{red!24}62.8 & \cellcolor{red!24}57.8 
    \\
    DEGREE-E2E
    & \cellcolor{red!00}70.3 & \cellcolor{red!00}66.8 & \cellcolor{red!00}55.1 & \cellcolor{red!00}49.1 
    & \cellcolor{red!00}67.7 & \cellcolor{red!00}60.5 & \cellcolor{red!00}48.7 & \cellcolor{red!00}43.7 
    & \cellcolor{red!00}74.7 & \cellcolor{red!00}70.2 & \cellcolor{red!00}33.8 & \cellcolor{red!00}23.3 
    & \cellcolor{red!00}61.6 & \cellcolor{red!00}59.2 & \cellcolor{red!00}35.6 & \cellcolor{red!00}25.4 
    \\
    DEGREE-PIPE
    & \cellcolor{red!00}72.0 & \cellcolor{red!00}68.4 & \cellcolor{red!08}56.3 & \cellcolor{red!00}50.7 
    & \cellcolor{red!00}68.3 & \cellcolor{red!08}61.7 & \cellcolor{red!08}48.9 & \cellcolor{red!08}44.8 
    & \cellcolor{red!00}74.0 & \cellcolor{red!00}70.4 & \cellcolor{red!08}49.6 & \cellcolor{red!08}42.7 
    & \cellcolor{red!00}63.7 & \cellcolor{red!00}60.5 & \cellcolor{red!00}49.3 & \cellcolor{red!08}39.8 
    \\

    \midrule
    \multirow{2}{*}{Model} & \multicolumn{4}{c|}{\textbf{Genia2013}} & \multicolumn{4}{c|}{\textbf{M$^2$E$^2$}} & \multicolumn{4}{c|}{\textbf{CASIE}} & \multicolumn{4}{c}{\textbf{PHEE}} \\
    & TI & TC & AC & AC+ & TI & TC & AC & AC+ & TI & TC & AC & AC+ & TI & TC & AC & AC+ \\
    \midrule
        DyGIE++
    & \cellcolor{red!08}76.3 & \cellcolor{red!08}72.9 & \cellcolor{red!16}60.5 & \cellcolor{red!16}57.2 
    & \cellcolor{red!16}53.1 & \cellcolor{red!16}51.0 & \cellcolor{red!00}33.4 & \cellcolor{red!08}30.8 
    & \cellcolor{red!00}44.9 & \cellcolor{red!00}44.7 & \cellcolor{red!08}36.4 & \cellcolor{red!08}29.5 
    & \cellcolor{red!08}71.4 & \cellcolor{red!08}70.4 & \cellcolor{red!24}60.8 & \cellcolor{red!24}45.7 
    \\
    OneIE
    & \cellcolor{red!16}78.0 & \cellcolor{red!16}74.3 & \cellcolor{red!08}51.0 & \cellcolor{red!00}32.9 
    & \cellcolor{red!08}52.4 & \cellcolor{red!08}50.6 & \cellcolor{red!24}36.1 & \cellcolor{red!16}32.1 
    & \cellcolor{red!16}70.8 & \cellcolor{red!16}70.6 & \cellcolor{red!16}54.2 & \cellcolor{red!00}22.1 
    & \cellcolor{red!00}70.9 & \cellcolor{red!00}70.0 & \cellcolor{red!00}37.5 & \cellcolor{red!00}29.8 
    \\
    AMR-IE
    & \cellcolor{red!24}78.0 & \cellcolor{red!24}74.5 & \cellcolor{red!00}34.8 & \cellcolor{red!00}23.1 
    & \cellcolor{red!08}52.4 & \cellcolor{red!08}50.5 & \cellcolor{red!08}35.5 & \cellcolor{red!08}31.9 
    & \cellcolor{red!24}71.1 & \cellcolor{red!24}70.8 & \cellcolor{red!00}10.7 & \cellcolor{red!00}3.1 
    & \cellcolor{red!00}70.2 & \cellcolor{red!00}69.4 & \cellcolor{red!00}45.7 & \cellcolor{red!00}34.1 
    \\
    EEQA
    & \cellcolor{red!00}72.4 & \cellcolor{red!00}69.4 & \cellcolor{red!00}48.1 & \cellcolor{red!08}35.7 
    & \cellcolor{red!24}53.6 & \cellcolor{red!24}51.0 & \cellcolor{red!00}32.6 & \cellcolor{red!00}30.2 
    & \cellcolor{red!00}43.2 & \cellcolor{red!00}42.8 & \cellcolor{red!00}35.1 & \cellcolor{red!08}26.2 
    & \cellcolor{red!08}70.9 & \cellcolor{red!08}70.3 & \cellcolor{red!00}40.4 & \cellcolor{red!00}32.0 
    \\
    RCEE
    & \cellcolor{red!00}71.4 & \cellcolor{red!00}68.0 & \cellcolor{red!00}45.8 & \cellcolor{red!00}31.6 
    & \cellcolor{red!00}50.1 & \cellcolor{red!00}48.1 & \cellcolor{red!00}31.0 & \cellcolor{red!00}28.0 
    & \cellcolor{red!00}42.3 & \cellcolor{red!00}42.1 & \cellcolor{red!00}32.8 & \cellcolor{red!00}23.7 
    & \cellcolor{red!16}71.6 & \cellcolor{red!16}70.9 & \cellcolor{red!00}41.6 & \cellcolor{red!00}33.1 
    \\
    Query\&Extract
    & \cellcolor{red!00}-- & \cellcolor{red!00}-- & \cellcolor{red!00}-- & \cellcolor{red!00}-- 
    & \cellcolor{red!00}51.4 & \cellcolor{red!00}49.4 & \cellcolor{red!08}33.9 & \cellcolor{red!00}28.8 
    & \cellcolor{red!00}-- & \cellcolor{red!00}-- & \cellcolor{red!00}-- & \cellcolor{red!00}-- 
    & \cellcolor{red!00}66.2 & \cellcolor{red!00}55.5 & \cellcolor{red!00}41.4 & \cellcolor{red!00}31.8 
    \\
    TagPrime
    & \cellcolor{red!08}75.7 & \cellcolor{red!08}73.0 & \cellcolor{red!24}60.8 & \cellcolor{red!24}57.4 
    & \cellcolor{red!00}52.2 & \cellcolor{red!00}50.2 & \cellcolor{red!16}35.5 & \cellcolor{red!24}32.4 
    & \cellcolor{red!08}69.5 & \cellcolor{red!08}69.3 & \cellcolor{red!24}61.0 & \cellcolor{red!24}49.1 
    & \cellcolor{red!24}71.7 & \cellcolor{red!24}71.1 & \cellcolor{red!16}51.7 & \cellcolor{red!16}40.6 
    \\
    DEGREE-E2E
    & \cellcolor{red!00}66.4 & \cellcolor{red!00}62.6 & \cellcolor{red!00}33.3 & \cellcolor{red!00}24.8 
    & \cellcolor{red!00}50.9 & \cellcolor{red!00}49.5 & \cellcolor{red!00}32.5 & \cellcolor{red!00}30.0 
    & \cellcolor{red!08}60.9 & \cellcolor{red!08}60.7 & \cellcolor{red!00}27.0 & \cellcolor{red!00}14.6 
    & \cellcolor{red!00}70.0 & \cellcolor{red!00}69.1 & \cellcolor{red!08}49.3 & \cellcolor{red!08}36.5 
    \\
    DEGREE-PIPE
    & \cellcolor{red!00}64.9 & \cellcolor{red!00}61.0 & \cellcolor{red!08}49.4 & \cellcolor{red!08}41.9 
    & \cellcolor{red!00}50.4 & \cellcolor{red!00}48.3 & \cellcolor{red!00}33.1 & \cellcolor{red!00}30.1 
    & \cellcolor{red!00}57.4 & \cellcolor{red!00}57.1 & \cellcolor{red!08}48.0 & \cellcolor{red!16}33.7 
    & \cellcolor{red!00}69.8 & \cellcolor{red!00}69.1 & \cellcolor{red!08}50.2 & \cellcolor{red!08}36.7 
    \\

    \bottomrule

\end{tabular}}
\caption{Reevaluation results for end-to-end event extraction (E2E). All the numbers are the average score of 5 data splits. Darker cells imply higher scores. We use ``--'' to denote the cases that models are not runnable.}
\label{tab:e2e}
\end{table*}

\begin{table*}[t!]
\centering
\setlength{\tabcolsep}{6pt}
\resizebox{.75\textwidth}{!}{
\begin{tabular}{l|cc|cc|cc|cc|cc|cc}
    \toprule
    \multirow{2}{*}{Model} & \multicolumn{2}{c|}{\textbf{ACE05}} & \multicolumn{2}{c|}{\textbf{RichERE}} & \multicolumn{2}{c|}{\textbf{MLEE}} & \multicolumn{2}{c|}{\textbf{Genia2011}} & \multicolumn{2}{c|}{\textbf{Genia2013}} & \multicolumn{2}{c}{\textbf{M$^2$E$^2$}} \\
    & TI & TC & TI & TC & TI & TC & TI & TC & TI & TC & TI & TC \\
    \midrule
    DyGIE++
    & \cellcolor{red!16}74.7 & \cellcolor{red!24}71.3 
    & \cellcolor{red!16}69.7 & \cellcolor{red!00}59.8 
    & \cellcolor{red!16}82.6 & \cellcolor{red!08}78.2 
    & \cellcolor{red!08}74.2 & \cellcolor{red!08}70.3 
    & \cellcolor{red!16}76.3 & \cellcolor{red!08}72.9 
    & \cellcolor{red!16}53.1 & \cellcolor{red!16}51.0 
    \\
    OneIE
    & \cellcolor{red!24}75.0 & \cellcolor{red!16}71.1 
    & \cellcolor{red!24}71.0 & \cellcolor{red!16}62.5 
    & \cellcolor{red!24}82.7 & \cellcolor{red!16}78.5 
    & \cellcolor{red!16}76.1 & \cellcolor{red!16}72.1 
    & \cellcolor{red!16}78.0 & \cellcolor{red!16}74.3 
    & \cellcolor{red!08}52.4 & \cellcolor{red!16}50.6 
    \\
    AMR-IE
    & \cellcolor{red!16}74.6 & \cellcolor{red!16}71.1 
    & \cellcolor{red!16}70.5 & \cellcolor{red!16}62.3 
    & \cellcolor{red!16}82.4 & \cellcolor{red!16}78.2 
    & \cellcolor{red!24}76.4 & \cellcolor{red!24}72.4 
    & \cellcolor{red!24}78.0 & \cellcolor{red!24}74.5 
    & \cellcolor{red!16}52.4 & \cellcolor{red!08}50.5 
    \\
    EEQA
    & \cellcolor{red!00}73.8 & \cellcolor{red!08}70.0 
    & \cellcolor{red!00}69.3 & \cellcolor{red!00}60.2 
    & \cellcolor{red!08}82.0 & \cellcolor{red!08}77.4 
    & \cellcolor{red!00}73.3 & \cellcolor{red!00}69.6 
    & \cellcolor{red!08}74.7 & \cellcolor{red!08}71.1 
    & \cellcolor{red!24}53.6 & \cellcolor{red!24}51.0 
    \\
    RCEE
    & \cellcolor{red!08}74.0 & \cellcolor{red!08}70.5 
    & \cellcolor{red!00}68.6 & \cellcolor{red!00}60.0 
    & \cellcolor{red!08}82.0 & \cellcolor{red!00}77.3 
    & \cellcolor{red!00}73.1 & \cellcolor{red!00}69.3 
    & \cellcolor{red!00}74.6 & \cellcolor{red!00}70.8 
    & \cellcolor{red!00}50.1 & \cellcolor{red!00}48.1 
    \\
    Query\&Extract
    & \cellcolor{red!00}68.6 & \cellcolor{red!00}65.1 
    & \cellcolor{red!00}67.5 & \cellcolor{red!00}59.8 
    & \cellcolor{red!00}78.0 & \cellcolor{red!00}74.9 
    & \cellcolor{red!00}71.6 & \cellcolor{red!00}68.9 
    & \cellcolor{red!00}73.0 & \cellcolor{red!00}70.1 
    & \cellcolor{red!00}51.4 & \cellcolor{red!00}49.4 
    \\
    TagPrime-C
    & \cellcolor{red!00}73.2 & \cellcolor{red!00}69.9 
    & \cellcolor{red!08}69.6 & \cellcolor{red!24}63.5 
    & \cellcolor{red!00}81.8 & \cellcolor{red!24}79.0 
    & \cellcolor{red!16}74.9 & \cellcolor{red!16}72.2 
    & \cellcolor{red!08}75.7 & \cellcolor{red!16}73.0 
    & \cellcolor{red!08}52.2 & \cellcolor{red!08}50.2 
    \\
    UniST
    & \cellcolor{red!08}73.9 & \cellcolor{red!00}69.8 
    & \cellcolor{red!08}69.6 & \cellcolor{red!08}60.7 
    & \cellcolor{red!00}80.2 & \cellcolor{red!00}74.9 
    & \cellcolor{red!08}73.8 & \cellcolor{red!08}70.3 
    & \cellcolor{red!00}73.7 & \cellcolor{red!00}69.9 
    & \cellcolor{red!00}51.1 & \cellcolor{red!00}49.0 
    \\
    CEDAR
    & \cellcolor{red!00}71.9 & \cellcolor{red!00}62.6 
    & \cellcolor{red!00}67.3 & \cellcolor{red!00}52.3 
    & \cellcolor{red!00}71.0 & \cellcolor{red!00}65.5 
    & \cellcolor{red!00}70.2 & \cellcolor{red!00}66.8 
    & \cellcolor{red!00}73.6 & \cellcolor{red!00}67.1 
    & \cellcolor{red!00}50.9 & \cellcolor{red!00}48.0 
    \\
    DEGREE
    & \cellcolor{red!00}72.0 & \cellcolor{red!00}68.4 
    & \cellcolor{red!00}68.3 & \cellcolor{red!08}61.7 
    & \cellcolor{red!00}74.0 & \cellcolor{red!00}70.4 
    & \cellcolor{red!00}63.7 & \cellcolor{red!00}60.5 
    & \cellcolor{red!00}64.9 & \cellcolor{red!00}61.0 
    & \cellcolor{red!00}50.4 & \cellcolor{red!00}48.3 
    \\

    \midrule
    \multirow{2}{*}{Model} & \multicolumn{2}{c|}{\textbf{CASIE}} & \multicolumn{2}{c|}{\textbf{PHEE}} & \multicolumn{2}{c|}{\textbf{MAVEN}} & \multicolumn{2}{c|}{\textbf{FewEvent}} & \multicolumn{2}{c|}{\textbf{MEE-en}} & \multicolumn{2}{c}{\textbf{SPEED}} \\
    & TI & TC & TI & TC & TI & TC & TI & TC & TI & TC & TI & TC \\
    \midrule
        DyGIE++
    & \cellcolor{red!00}44.9 & \cellcolor{red!00}44.7 
    & \cellcolor{red!16}71.4 & \cellcolor{red!16}70.4 
    & \cellcolor{red!08}75.9 & \cellcolor{red!08}65.3 
    & \cellcolor{red!16}67.7 & \cellcolor{red!08}65.2 
    & \cellcolor{red!24}81.7 & \cellcolor{red!16}79.8 
    & \cellcolor{red!08}69.6 & \cellcolor{red!00}64.9 
    \\
    OneIE
    & \cellcolor{red!16}70.8 & \cellcolor{red!16}70.6 
    & \cellcolor{red!00}70.9 & \cellcolor{red!00}70.0 
    & \cellcolor{red!16}76.4 & \cellcolor{red!16}65.5 
    & \cellcolor{red!08}67.5 & \cellcolor{red!16}65.4 
    & \cellcolor{red!00}80.7 & \cellcolor{red!08}78.8 
    & \cellcolor{red!00}69.5 & \cellcolor{red!08}65.1 
    \\
    AMR-IE
    & \cellcolor{red!24}71.1 & \cellcolor{red!24}70.8 
    & \cellcolor{red!00}70.2 & \cellcolor{red!00}69.4 
    & \cellcolor{red!00}-- & \cellcolor{red!00}-- 
    & \cellcolor{red!08}67.4 & \cellcolor{red!08}65.2 
    & \cellcolor{red!00}-- & \cellcolor{red!00}-- 
    & \cellcolor{red!00}-- & \cellcolor{red!00}-- 
    \\
    EEQA
    & \cellcolor{red!00}43.4 & \cellcolor{red!00}43.2 
    & \cellcolor{red!08}70.9 & \cellcolor{red!08}70.3 
    & \cellcolor{red!00}75.2 & \cellcolor{red!00}64.4 
    & \cellcolor{red!00}67.0 & \cellcolor{red!00}65.1 
    & \cellcolor{red!08}81.4 & \cellcolor{red!16}79.5 
    & \cellcolor{red!08}69.9 & \cellcolor{red!16}65.3 
    \\
    RCEE
    & \cellcolor{red!00}43.5 & \cellcolor{red!00}43.3 
    & \cellcolor{red!16}71.6 & \cellcolor{red!16}70.9 
    & \cellcolor{red!00}75.2 & \cellcolor{red!08}64.6 
    & \cellcolor{red!00}67.0 & \cellcolor{red!00}65.0 
    & \cellcolor{red!08}81.1 & \cellcolor{red!08}79.1 
    & \cellcolor{red!16}70.1 & \cellcolor{red!08}65.1 
    \\
    Query\&Extract
    & \cellcolor{red!00}51.6 & \cellcolor{red!00}51.5 
    & \cellcolor{red!00}66.2 & \cellcolor{red!00}55.5 
    & \cellcolor{red!00}-- & \cellcolor{red!00}-- 
    & \cellcolor{red!00}66.3 & \cellcolor{red!00}63.8 
    & \cellcolor{red!00}80.2 & \cellcolor{red!00}78.1 
    & \cellcolor{red!16}70.2 & \cellcolor{red!16}66.2 
    \\
    TagPrime-C
    & \cellcolor{red!16}69.5 & \cellcolor{red!16}69.3 
    & \cellcolor{red!24}71.7 & \cellcolor{red!24}71.1 
    & \cellcolor{red!00}74.7 & \cellcolor{red!24}66.1 
    & \cellcolor{red!00}67.2 & \cellcolor{red!24}65.6 
    & \cellcolor{red!16}81.5 & \cellcolor{red!24}79.8 
    & \cellcolor{red!24}70.3 & \cellcolor{red!24}66.4 
    \\
    UniST
    & \cellcolor{red!08}68.4 & \cellcolor{red!08}68.1 
    & \cellcolor{red!00}70.7 & \cellcolor{red!00}69.6 
    & \cellcolor{red!24}76.7 & \cellcolor{red!00}63.4 
    & \cellcolor{red!16}67.5 & \cellcolor{red!00}63.1 
    & \cellcolor{red!00}80.5 & \cellcolor{red!00}78.3 
    & \cellcolor{red!00}-- & \cellcolor{red!00}-- 
    \\
    CEDAR
    & \cellcolor{red!08}68.7 & \cellcolor{red!08}67.6 
    & \cellcolor{red!08}71.2 & \cellcolor{red!08}70.3 
    & \cellcolor{red!16}76.5 & \cellcolor{red!00}54.5 
    & \cellcolor{red!00}66.9 & \cellcolor{red!00}52.1 
    & \cellcolor{red!16}81.5 & \cellcolor{red!00}78.6 
    & \cellcolor{red!00}67.6 & \cellcolor{red!00}61.7 
    \\
    DEGREE
    & \cellcolor{red!00}61.5 & \cellcolor{red!00}61.3 
    & \cellcolor{red!00}69.8 & \cellcolor{red!00}69.1 
    & \cellcolor{red!08}76.2 & \cellcolor{red!16}65.5 
    & \cellcolor{red!24}67.9 & \cellcolor{red!16}65.5 
    & \cellcolor{red!00}80.2 & \cellcolor{red!00}78.2 
    & \cellcolor{red!00}66.5 & \cellcolor{red!00}62.2 
    \\
    \bottomrule

\end{tabular}}
\caption{Reevaluation results for event detection (ED). All the numbers are the average score of 5 data splits. Darker cells imply higher scores. We use ``--'' to denote the cases that models are not runnable.}
\label{tab:ed}
\vspace{-1em}
\end{table*}

\begin{table*}[t!]
\centering
\setlength{\tabcolsep}{5.5pt}
\resizebox{.99\textwidth}{!}{
\begin{tabular}{l|ccc|ccc|ccc|ccc|ccc|ccc}
    \toprule
    \multirow{2}{*}{Model} & \multicolumn{3}{c|}{\textbf{ACE05}} & \multicolumn{3}{c|}{\textbf{RichERE}} & \multicolumn{3}{c|}{\textbf{MLEE}} & \multicolumn{3}{c|}{\textbf{Genia2011}} & \multicolumn{3}{c|}{\textbf{Genia2013}} & \multicolumn{3}{c}{\textbf{M$^2$E$^2$}} \\
    & AI & AC & AC+ & AI & AC & AC+ & AI & AC & AC+ & AI & AC & AC+ & AI & AC & AC+ & AI & AC & AC+ \\
    \midrule
    DyGIE++
    & \cellcolor{red!00}66.9 & \cellcolor{red!00}61.5 & \cellcolor{red!00}60.0 
    & \cellcolor{red!00}58.5 & \cellcolor{red!00}49.4 & \cellcolor{red!00}47.3 
    & \cellcolor{red!08}67.9 & \cellcolor{red!00}64.8 & \cellcolor{red!08}62.4 
    & \cellcolor{red!00}66.1 & \cellcolor{red!00}63.7 & \cellcolor{red!00}61.0 
    & \cellcolor{red!08}71.7 & \cellcolor{red!08}69.3 & \cellcolor{red!08}66.9 
    & \cellcolor{red!00}41.7 & \cellcolor{red!00}38.9 & \cellcolor{red!00}38.5 
    \\
    OneIE
    & \cellcolor{red!00}75.4 & \cellcolor{red!00}71.5 & \cellcolor{red!00}70.2 
    & \cellcolor{red!00}71.6 & \cellcolor{red!00}65.8 & \cellcolor{red!00}63.7 
    & \cellcolor{red!00}31.0 & \cellcolor{red!00}28.9 & \cellcolor{red!00}15.7 
    & \cellcolor{red!00}62.9 & \cellcolor{red!00}60.3 & \cellcolor{red!00}38.9 
    & \cellcolor{red!00}57.2 & \cellcolor{red!00}55.7 & \cellcolor{red!00}38.7 
    & \cellcolor{red!00}59.0 & \cellcolor{red!00}55.2 & \cellcolor{red!00}53.3 
    \\
    AMR-IE
    & \cellcolor{red!08}76.2 & \cellcolor{red!00}72.6 & \cellcolor{red!00}70.9 
    & \cellcolor{red!00}72.8 & \cellcolor{red!00}65.8 & \cellcolor{red!00}63.0 
    & \cellcolor{red!00}23.2 & \cellcolor{red!00}16.6 & \cellcolor{red!00}6.1 
    & \cellcolor{red!00}49.1 & \cellcolor{red!00}47.6 & \cellcolor{red!00}35.3 
    & \cellcolor{red!00}38.9 & \cellcolor{red!00}38.1 & \cellcolor{red!00}26.4 
    & \cellcolor{red!00}56.0 & \cellcolor{red!00}51.3 & \cellcolor{red!00}50.4 
    \\
    EEQA
    & \cellcolor{red!00}73.8 & \cellcolor{red!00}71.4 & \cellcolor{red!00}69.6 
    & \cellcolor{red!00}73.3 & \cellcolor{red!00}67.3 & \cellcolor{red!00}64.9 
    & \cellcolor{red!00}64.8 & \cellcolor{red!00}62.1 & \cellcolor{red!00}49.5 
    & \cellcolor{red!00}63.2 & \cellcolor{red!00}60.8 & \cellcolor{red!00}49.4 
    & \cellcolor{red!00}64.7 & \cellcolor{red!00}61.1 & \cellcolor{red!00}47.5 
    & \cellcolor{red!00}57.6 & \cellcolor{red!00}55.9 & \cellcolor{red!00}55.3 
    \\
    RCEE
    & \cellcolor{red!00}73.7 & \cellcolor{red!00}71.2 & \cellcolor{red!00}69.4 
    & \cellcolor{red!00}72.8 & \cellcolor{red!00}67.0 & \cellcolor{red!00}64.5 
    & \cellcolor{red!00}61.1 & \cellcolor{red!00}58.2 & \cellcolor{red!00}45.1 
    & \cellcolor{red!00}62.3 & \cellcolor{red!00}59.9 & \cellcolor{red!00}49.6 
    & \cellcolor{red!00}60.7 & \cellcolor{red!00}57.4 & \cellcolor{red!00}42.7 
    & \cellcolor{red!00}57.9 & \cellcolor{red!00}56.4 & \cellcolor{red!00}55.8 
    \\
    Query\&Extract
    & \cellcolor{red!16}77.3 & \cellcolor{red!16}73.6 & \cellcolor{red!16}72.0 
    & \cellcolor{red!16}76.4 & \cellcolor{red!16}70.9 & \cellcolor{red!16}69.2 
    & \cellcolor{red!00}-- & \cellcolor{red!00}-- & \cellcolor{red!00}-- 
    & \cellcolor{red!00}-- & \cellcolor{red!00}-- & \cellcolor{red!00}-- 
    & \cellcolor{red!00}-- & \cellcolor{red!00}-- & \cellcolor{red!00}-- 
    & \cellcolor{red!00}59.9 & \cellcolor{red!00}56.2 & \cellcolor{red!00}54.2 
    \\
    TagPrime-C
    & \cellcolor{red!24}80.0 & \cellcolor{red!24}76.0 & \cellcolor{red!24}74.5 
    & \cellcolor{red!24}78.8 & \cellcolor{red!24}73.3 & \cellcolor{red!24}71.4 
    & \cellcolor{red!24}78.9 & \cellcolor{red!24}76.6 & \cellcolor{red!24}74.5 
    & \cellcolor{red!24}79.6 & \cellcolor{red!24}77.4 & \cellcolor{red!24}75.8 
    & \cellcolor{red!24}79.8 & \cellcolor{red!24}77.4 & \cellcolor{red!24}74.9 
    & \cellcolor{red!24}63.4 & \cellcolor{red!16}60.1 & \cellcolor{red!08}59.0 
    \\
    TagPrime-CR
    & \cellcolor{red!24}80.1 & \cellcolor{red!24}77.8 & \cellcolor{red!24}76.2 
    & \cellcolor{red!24}78.7 & \cellcolor{red!24}74.3 & \cellcolor{red!24}72.5 
    & \cellcolor{red!24}79.2 & \cellcolor{red!24}77.3 & \cellcolor{red!24}74.6 
    & \cellcolor{red!24}78.0 & \cellcolor{red!24}76.2 & \cellcolor{red!24}74.5 
    & \cellcolor{red!16}76.6 & \cellcolor{red!16}74.5 & \cellcolor{red!16}72.3 
    & \cellcolor{red!24}63.2 & \cellcolor{red!24}60.8 & \cellcolor{red!24}59.9 
    \\
    DEGREE
    & \cellcolor{red!08}76.4 & \cellcolor{red!08}73.3 & \cellcolor{red!08}71.8 
    & \cellcolor{red!08}75.1 & \cellcolor{red!08}70.2 & \cellcolor{red!08}68.8 
    & \cellcolor{red!00}67.6 & \cellcolor{red!08}65.3 & \cellcolor{red!00}61.5 
    & \cellcolor{red!00}68.2 & \cellcolor{red!00}65.7 & \cellcolor{red!00}62.4 
    & \cellcolor{red!00}68.4 & \cellcolor{red!00}66.0 & \cellcolor{red!00}62.5 
    & \cellcolor{red!08}62.3 & \cellcolor{red!08}59.8 & \cellcolor{red!16}59.2 
    \\
    BART-Gen
    & \cellcolor{red!00}76.0 & \cellcolor{red!08}72.6 & \cellcolor{red!08}71.2 
    & \cellcolor{red!00}74.4 & \cellcolor{red!00}68.8 & \cellcolor{red!00}67.7 
    & \cellcolor{red!16}73.1 & \cellcolor{red!16}69.8 & \cellcolor{red!16}68.7 
    & \cellcolor{red!16}73.4 & \cellcolor{red!16}70.9 & \cellcolor{red!16}69.5 
    & \cellcolor{red!16}76.4 & \cellcolor{red!16}73.6 & \cellcolor{red!16}72.2 
    & \cellcolor{red!08}62.5 & \cellcolor{red!16}60.0 & \cellcolor{red!16}59.6 
    \\
    X-Gear
    & \cellcolor{red!00}76.1 & \cellcolor{red!00}72.4 & \cellcolor{red!00}70.8 
    & \cellcolor{red!08}75.0 & \cellcolor{red!00}68.7 & \cellcolor{red!00}67.2 
    & \cellcolor{red!00}64.8 & \cellcolor{red!00}63.3 & \cellcolor{red!00}59.4 
    & \cellcolor{red!08}68.4 & \cellcolor{red!08}66.2 & \cellcolor{red!08}63.1 
    & \cellcolor{red!00}64.1 & \cellcolor{red!00}61.9 & \cellcolor{red!00}58.6 
    & \cellcolor{red!16}62.7 & \cellcolor{red!08}59.8 & \cellcolor{red!08}59.0 
    \\
    PAIE
    & \cellcolor{red!16}77.2 & \cellcolor{red!16}74.0 & \cellcolor{red!16}72.9 
    & \cellcolor{red!16}76.6 & \cellcolor{red!16}71.1 & \cellcolor{red!16}70.0 
    & \cellcolor{red!16}76.0 & \cellcolor{red!16}73.5 & \cellcolor{red!16}72.4 
    & \cellcolor{red!16}76.8 & \cellcolor{red!16}74.6 & \cellcolor{red!16}73.4 
    & \cellcolor{red!24}77.8 & \cellcolor{red!24}75.2 & \cellcolor{red!24}74.2 
    & \cellcolor{red!16}62.9 & \cellcolor{red!24}60.6 & \cellcolor{red!24}60.4 
    \\
    Ampere
    & \cellcolor{red!00}75.5 & \cellcolor{red!00}72.0 & \cellcolor{red!00}70.6 
    & \cellcolor{red!00}73.8 & \cellcolor{red!08}69.2 & \cellcolor{red!08}67.7 
    & \cellcolor{red!08}69.2 & \cellcolor{red!08}67.1 & \cellcolor{red!08}62.6 
    & \cellcolor{red!08}69.5 & \cellcolor{red!08}67.1 & \cellcolor{red!08}63.8 
    & \cellcolor{red!08}73.2 & \cellcolor{red!08}71.0 & \cellcolor{red!08}67.7 
    & \cellcolor{red!00}62.1 & \cellcolor{red!00}59.1 & \cellcolor{red!00}58.4 
    \\

    \midrule
    \multirow{2}{*}{Model} & \multicolumn{3}{c|}{\textbf{CASIE}} & \multicolumn{3}{c|}{\textbf{PHEE}} & \multicolumn{3}{c|}{\textbf{WikiEvents}} & \multicolumn{3}{c|}{\textbf{RAMS}} & \multicolumn{3}{c|}{\textbf{GENEVA}} & \multicolumn{3}{c}{\textbf{MUC-4}} \\
    & AI & AC & AC+ & AI & AC & AC+ & AI & AC & AC+ & AI & AC & AC+ & AI & AC & AC+ & AI & AC & AC+ \\
    \midrule
    DyGIE++
    & \cellcolor{red!00}58.0 & \cellcolor{red!00}56.0 & \cellcolor{red!00}51.5 
    & \cellcolor{red!08}63.4 & \cellcolor{red!08}54.6 & \cellcolor{red!08}54.2 
    & \cellcolor{red!00}39.8 & \cellcolor{red!00}35.3 & \cellcolor{red!00}34.7 
    & \cellcolor{red!00}44.3 & \cellcolor{red!00}35.3 & \cellcolor{red!00}35.3 
    & \cellcolor{red!00}66.0 & \cellcolor{red!00}62.5 & \cellcolor{red!00}62.3 
    & \cellcolor{red!24}56.5 & \cellcolor{red!24}55.6 & \cellcolor{red!24}55.6 
    \\
    OneIE
    & \cellcolor{red!00}58.3 & \cellcolor{red!00}55.3 & \cellcolor{red!00}27.7 
    & \cellcolor{red!00}55.9 & \cellcolor{red!00}40.6 & \cellcolor{red!00}40.4 
    & \cellcolor{red!00}17.5 & \cellcolor{red!00}15.0 & \cellcolor{red!00}7.9 
    & \cellcolor{red!00}48.0 & \cellcolor{red!00}40.7 & \cellcolor{red!00}40.7 
    & \cellcolor{red!00}38.9 & \cellcolor{red!00}37.1 & \cellcolor{red!00}36.9 
    & \cellcolor{red!16}55.1 & \cellcolor{red!16}53.9 & \cellcolor{red!16}53.9 
    \\
    AMR-IE
    & \cellcolor{red!00}35.5 & \cellcolor{red!00}11.0 & \cellcolor{red!00}4.0 
    & \cellcolor{red!00}60.4 & \cellcolor{red!00}45.3 & \cellcolor{red!00}44.9 
    & \cellcolor{red!00}17.8 & \cellcolor{red!00}16.0 & \cellcolor{red!00}10.4 
    & \cellcolor{red!00}49.6 & \cellcolor{red!00}42.3 & \cellcolor{red!00}42.3 
    & \cellcolor{red!00}23.7 & \cellcolor{red!00}16.6 & \cellcolor{red!00}16.4 
    & \cellcolor{red!00}-- & \cellcolor{red!00}-- & \cellcolor{red!00}-- 
    \\
    EEQA
    & \cellcolor{red!00}56.1 & \cellcolor{red!00}54.0 & \cellcolor{red!00}49.0 
    & \cellcolor{red!00}53.7 & \cellcolor{red!00}45.6 & \cellcolor{red!00}45.4 
    & \cellcolor{red!00}54.3 & \cellcolor{red!00}51.7 & \cellcolor{red!00}46.1 
    & \cellcolor{red!00}48.9 & \cellcolor{red!00}44.7 & \cellcolor{red!00}44.7 
    & \cellcolor{red!08}69.7 & \cellcolor{red!08}67.3 & \cellcolor{red!08}67.0 
    & \cellcolor{red!00}32.7 & \cellcolor{red!00}27.4 & \cellcolor{red!00}27.4 
    \\
    RCEE
    & \cellcolor{red!00}47.6 & \cellcolor{red!00}45.3 & \cellcolor{red!00}39.5 
    & \cellcolor{red!00}54.1 & \cellcolor{red!00}45.8 & \cellcolor{red!00}45.6 
    & \cellcolor{red!00}53.7 & \cellcolor{red!00}50.9 & \cellcolor{red!00}44.0 
    & \cellcolor{red!00}45.4 & \cellcolor{red!00}41.5 & \cellcolor{red!00}41.5 
    & \cellcolor{red!00}66.2 & \cellcolor{red!00}63.8 & \cellcolor{red!00}63.4 
    & \cellcolor{red!00}33.0 & \cellcolor{red!00}28.1 & \cellcolor{red!00}28.1 
    \\
    Query\&Extract
    & \cellcolor{red!00}-- & \cellcolor{red!00}-- & \cellcolor{red!00}-- 
    & \cellcolor{red!08}64.6 & \cellcolor{red!08}54.8 & \cellcolor{red!08}54.4 
    & \cellcolor{red!00}-- & \cellcolor{red!00}-- & \cellcolor{red!00}-- 
    & \cellcolor{red!00}-- & \cellcolor{red!00}-- & \cellcolor{red!00}-- 
    & \cellcolor{red!00}52.2 & \cellcolor{red!00}50.3 & \cellcolor{red!00}50.0 
    & \cellcolor{red!00}-- & \cellcolor{red!00}-- & \cellcolor{red!00}-- 
    \\
    TagPrime-C
    & \cellcolor{red!24}71.9 & \cellcolor{red!24}69.1 & \cellcolor{red!24}66.1 
    & \cellcolor{red!16}66.0 & \cellcolor{red!16}55.6 & \cellcolor{red!16}55.3 
    & \cellcolor{red!24}70.4 & \cellcolor{red!24}65.7 & \cellcolor{red!16}64.0 
    & \cellcolor{red!24}54.4 & \cellcolor{red!16}48.3 & \cellcolor{red!16}48.3 
    & \cellcolor{red!24}83.0 & \cellcolor{red!24}79.2 & \cellcolor{red!24}79.0 
    & \cellcolor{red!16}55.3 & \cellcolor{red!16}54.4 & \cellcolor{red!16}54.4 
    \\
    TagPrime-CR
    & \cellcolor{red!24}71.1 & \cellcolor{red!24}69.2 & \cellcolor{red!24}66.1 
    & \cellcolor{red!16}65.8 & \cellcolor{red!16}56.0 & \cellcolor{red!16}55.7 
    & \cellcolor{red!24}70.3 & \cellcolor{red!24}67.2 & \cellcolor{red!24}65.5 
    & \cellcolor{red!16}54.1 & \cellcolor{red!24}49.7 & \cellcolor{red!24}49.7 
    & \cellcolor{red!24}82.8 & \cellcolor{red!24}80.4 & \cellcolor{red!24}80.1 
    & \cellcolor{red!24}55.5 & \cellcolor{red!24}54.7 & \cellcolor{red!24}54.7 
    \\
    DEGREE
    & \cellcolor{red!00}61.0 & \cellcolor{red!08}59.0 & \cellcolor{red!08}54.7 
    & \cellcolor{red!00}61.7 & \cellcolor{red!00}52.5 & \cellcolor{red!00}52.3 
    & \cellcolor{red!08}60.4 & \cellcolor{red!08}57.3 & \cellcolor{red!08}53.9 
    & \cellcolor{red!08}50.5 & \cellcolor{red!08}45.5 & \cellcolor{red!08}45.5 
    & \cellcolor{red!00}67.2 & \cellcolor{red!00}64.1 & \cellcolor{red!00}63.9 
    & \cellcolor{red!08}52.5 & \cellcolor{red!08}51.5 & \cellcolor{red!08}51.5 
    \\
    BART-Gen
    & \cellcolor{red!08}63.7 & \cellcolor{red!08}60.0 & \cellcolor{red!08}58.3 
    & \cellcolor{red!00}57.1 & \cellcolor{red!00}47.7 & \cellcolor{red!00}47.5 
    & \cellcolor{red!16}68.5 & \cellcolor{red!16}64.2 & \cellcolor{red!16}63.9 
    & \cellcolor{red!00}50.4 & \cellcolor{red!00}45.4 & \cellcolor{red!00}45.4 
    & \cellcolor{red!00}67.3 & \cellcolor{red!00}64.4 & \cellcolor{red!00}64.3 
    & \cellcolor{red!00}51.3 & \cellcolor{red!00}49.8 & \cellcolor{red!00}49.8 
    \\
    X-Gear
    & \cellcolor{red!16}65.7 & \cellcolor{red!16}63.4 & \cellcolor{red!16}59.3 
    & \cellcolor{red!24}67.6 & \cellcolor{red!24}58.3 & \cellcolor{red!24}58.2 
    & \cellcolor{red!00}58.7 & \cellcolor{red!00}55.6 & \cellcolor{red!00}52.4 
    & \cellcolor{red!16}52.1 & \cellcolor{red!08}46.2 & \cellcolor{red!08}46.2 
    & \cellcolor{red!16}78.9 & \cellcolor{red!16}75.1 & \cellcolor{red!16}74.9 
    & \cellcolor{red!08}51.5 & \cellcolor{red!08}50.4 & \cellcolor{red!08}50.4 
    \\
    PAIE
    & \cellcolor{red!16}68.1 & \cellcolor{red!16}65.7 & \cellcolor{red!16}64.0 
    & \cellcolor{red!24}74.9 & \cellcolor{red!24}73.3 & \cellcolor{red!24}73.1 
    & \cellcolor{red!16}69.8 & \cellcolor{red!16}65.5 & \cellcolor{red!24}65.2 
    & \cellcolor{red!24}55.2 & \cellcolor{red!24}50.5 & \cellcolor{red!24}50.5 
    & \cellcolor{red!16}73.5 & \cellcolor{red!16}70.4 & \cellcolor{red!16}70.3 
    & \cellcolor{red!00}48.8 & \cellcolor{red!00}47.9 & \cellcolor{red!00}47.9 
    \\
    Ampere
    & \cellcolor{red!08}61.1 & \cellcolor{red!00}58.4 & \cellcolor{red!00}53.9 
    & \cellcolor{red!00}61.4 & \cellcolor{red!00}51.7 & \cellcolor{red!00}51.6 
    & \cellcolor{red!08}59.9 & \cellcolor{red!08}56.7 & \cellcolor{red!08}53.3 
    & \cellcolor{red!08}52.0 & \cellcolor{red!16}46.8 & \cellcolor{red!16}46.8 
    & \cellcolor{red!08}67.8 & \cellcolor{red!08}65.0 & \cellcolor{red!08}64.8 
    & \cellcolor{red!00}-- & \cellcolor{red!00}-- & \cellcolor{red!00}-- 
    \\

    \bottomrule

\end{tabular}}
\caption{Reevaluation results for event argument extraction (EAE). All the numbers are the average score of 5 data splits. Darker cells imply higher scores. We use ``--'' to denote the cases that models are not runnable.}
\label{tab:eae}
\vspace{-1em}
\end{table*}

\mypar{Models.}
We do our best to aggregate as many models as possible into \mytool{}.
For those works having public codebases, we adapt their code to fit our evaluation framework.
We also re-implement some models based on the description from the original papers.
Currently, \mytool{} supports the following models:
(1) \emph{Joint training models} that train ED and EAE together in an end-to-end manner, including \textbf{DyGIE} \cite{Wadden19dygiepp}, \textbf{OneIE} \cite{Lin20oneie}, and \textbf{AMR-IE} \cite{Zhang21amrie}.
(2) \emph{Classification-based models} that formulate the event extraction task as a token classification problem, a sequential labeling problem, or a question answering problem, including \textbf{EEQA} \cite{Du20bertqa}, \textbf{RCEE} \cite{Liu20rcee}, \textbf{Query\&Extract} \cite{Wang22queryextract}, \textbf{TagPrime} \cite{Hsu22tagprime}, \textbf{UniST} \cite{Huang22unist}, and \textbf{CEDAR} \cite{Li23glen}.
(3) \emph{Generation-based models} that convert the event extraction task to a conditional generation problem, including \textbf{DEGREE} \cite{Hsu22degree}, \textbf{BART-Gen} \cite{Li21wikievents}, \textbf{X-Gear} \cite{Huang22xgear}, \textbf{PAIE} \cite{Ma22paie}, and \textbf{AMPERE} \cite{Hsu23ampere}.
Table~\ref{tab:models} presents the different assumptions and requirements for each model.
It is worth noting that some models need additional annotations or information, as indicated in the table.
Appendix~\ref{app:models} lists more details about implementations.

\mypar{Reevalutation results.}
For a fair comparison, we utilize RoBERTa-large \cite{Liu19roberta} for all the classification-based models and use BART-large \cite{Lewis20bart} for all the generation-based models.
Table~\ref{tab:e2e}, \ref{tab:ed}, and \ref{tab:eae} present the reevaluation results of end-to-end EE, ED, and EAE, respectively.
Appendix~\ref{app:results} lists more detailed results.
We first notice that for end-to-end EE and ED, there is no obvious dominant approach.
It suggests that the reported improvements from previous studies may be influenced by dataset bias, data split bias, or data processing.
This verifies the importance of a comprehensive evaluation framework that covers various domains of datasets and standardized data splits.
TagPrime \cite{Hsu22tagprime} and PAIE \cite{Ma22paie} seem to be the two dominant approaches across different types of datasets for EAE.
These results validate the effectiveness of those two models, aligning with our expectations for guiding reliable and reproducible research in event extraction with \mytool{}.

In addition, we observe a gap between the established evaluation metrics (AI and AC) and the proposed ones (AI+ and AC+).
This implies a potential mismatch between the earlier metrics and the predictive quality. 
We strongly recommend reporting the attaching score (AI+ and AC+) for future research in event extraction to provide a more accurate assessment of performance.

%% file: 05-llm.tex
\section{Have LLMs Solved Event Extraction?}
\label{sec:llm}

Given the demonstrated potential of large language models (LLMs) across various NLP tasks, we discuss their capability in solving event extraction tasks.
In contrast to previous studies \cite{Li23gpteval,Gao23chatgptee}, which evaluate a \emph{single} LLM on a \emph{single} EE dataset, we investigate multiple popular LLMs across multiple datasets provided by \mytool{}.
We consider \textbf{GPT-3.5-Turbo} as well as some open-source LLMs that achieve strong performance on Chatbot Arena \cite{Zheng23chatbotarena}\footnote{\url{https://leaderboard.lmsys.org}}, including \textbf{Llama-2-13b-chat-hf} and \textbf{Llama-2-70b-chat-hf} \cite{touvron2023llama}, \textbf{Zephyr-7b-alpha} \cite{Tunstall23zephyr}, and \textbf{Mixtral-8x7B-Instruct} \cite{Jiang24mixtral}, with vLLM framework \cite{Kwon23vllm}.
We evaluate them on the pipelined tasks of event detection (ED) and event argument extraction (EAE).
As part of the prompt, we provide LLMs with the task instructions, a few demonstration examples (positive and negative ones), and the query text.
It is worth noting that the number of demonstration examples will be limited by the maximum length supported by LLMs. 
Appendix~\ref{app:prompts} illustrates the best prompt we use.

\begin{table}[t!]
\centering
\setlength{\tabcolsep}{7pt}
\resizebox{0.9\columnwidth}{!}{
\begin{tabular}{l|cc}
    \toprule
    \textbf{Model} & \textbf{TI} & \textbf{TC}  \\
    \midrule
    OneIE \cite{Lin20oneie}                     & 73.5 & 69.5 \\
    TagPrime-C \cite{Hsu22tagprime}             & 72.5 & 69.5 \\
    \midrule 
    Llama-2-13b-chat-hf \emph{(2-shot)}         & 23.5 & 9.3 \\
    Llama-2-13b-chat-hf \emph{(6-shot)}         & 28.0 & 10.4 \\
    \midrule 
    Llama-2-70b-chat-hf \emph{(2-shot)}         & 30.6 & 11.3 \\
    Llama-2-70b-chat-hf \emph{(6-shot)}         & 32.2 & 12.4 \\
    \midrule 
    Zephyr-7b-alpha \emph{(2-shot)}             & 25.0 & 6.6 \\
    Zephyr-7b-alpha \emph{(6-shot)}             & 26.1 & 8.0 \\
    Zephyr-7b-alpha \emph{(16-shot)}            & 26.1 & 9.1 \\
    Zephyr-7b-alpha \emph{(32-shot)}            & 25.2 & 10.1 \\
    Zephyr-7b-alpha \emph{(64-shot)}            & 23.8 & 9.7 \\
    \midrule
    Mixtral-8x7B-Instruct-v0.1 \emph{(2-shot)}  & 30.4 & 10.2 \\
    Mixtral-8x7B-Instruct-v0.1 \emph{(6-shot)}  & 34.4 & 10.6 \\
    Mixtral-8x7B-Instruct-v0.1 \emph{(16-shot)} & 35.4 & 12.1 \\
    Mixtral-8x7B-Instruct-v0.1 \emph{(32-shot)} & 36.7 & 13.8 \\
    Mixtral-8x7B-Instruct-v0.1 \emph{(64-shot)} & 37.5 & 14.6 \\
    \midrule
    gpt-3.5-turbo-1106 \emph{(2-shot)}          & 33.9 & 11.8 \\
    gpt-3.5-turbo-1106 \emph{(16-shot)}         & 35.2 & 12.3 \\
    \bottomrule

\end{tabular}}
\caption{Average results over all datasets for event detection (ED) on sampled 250 documents.}
\label{tab:llm_ed}
\vspace{-1em}
\end{table}

\mypar{Results.}
Due to the cost and time of running LLMs, we evaluate only on sampled 250 documents for each dataset.
Table~\ref{tab:llm_ed} and \ref{tab:llm_eae} list the average results of LLMs as well as some well-performed models selected from \mytool{}.\footnote{The results do not include SPEED and MUC-4.} 
Unlike other NLP tasks such as named entity recognition and commonsense knowledge, where LLMs can achieve competitive performance with fine-tuning models using only a few in-context demonstrations \cite{Wei22llmgood1,Qin23llmgood2}, it is noteworthy that there is a large gap between LLMs and the baselines for both the ED and EAE tasks.
Our hypothesis is that event extraction requires more recognition of abstract concepts and relations, which is harder compared to other NLP tasks \cite{Li23gpteval}.

\begin{table}[t!]
\centering
\setlength{\tabcolsep}{5pt}
\resizebox{\columnwidth}{!}{
\begin{tabular}{l|cccc}
    \toprule
    \textbf{Model} & \textbf{AI} & \textbf{AC} & \textbf{AI+} & \textbf{AC+} \\
    \midrule
    TagPrime-CR \cite{Hsu22tagprime}            & 73.3 & 69.5 & 71.9 & 68.1 \\
    PAIE \cite{Ma22paie}                        & 72.0 & 68.9 & 71.3 & 68.1 \\
    \midrule
    Llama-2-13b-chat-hf \emph{(2-shot)}         & 26.5 & 19.0 & 24.1 & 17.1 \\
    Llama-2-13b-chat-hf \emph{(4-shot)}         & 25.0 & 18.7 & 22.8 & 17.0 \\
    \midrule
    Llama-2-70b-chat-hf \emph{(2-shot)}         & 30.6 & 24.4 & 28.5 & 22.8 \\
    Llama-2-70b-chat-hf \emph{(4-shot)}         & 30.1 & 23.6 & 28.3 & 22.3 \\
    \midrule
    Zephyr-7b-alpha \emph{(2-shot)}             & 28.9 & 22.6 & 27.0 & 21.3 \\
    Zephyr-7b-alpha \emph{(4-shot)}             & 29.3 & 23.9 & 27.0 & 22.4 \\
    Zephyr-7b-alpha \emph{(8-shot)}             & 29.7 & 25.2 & 27.7 & 23.5 \\
    Zephyr-7b-alpha \emph{(16-shot)}            & 27.2 & 22.5 & 26.3 & 21.8 \\
    Zephyr-7b-alpha \emph{(32-shot)}            & 24.3 & 19.7 & 23.7 & 19.3 \\
    \midrule
    Mixtral-8x7B-Instruct-v0.1 \emph{(2-shot)}  & 28.5 & 23.6 & 26.7 & 22.2 \\
    Mixtral-8x7B-Instruct-v0.1 \emph{(4-shot)}  & 30.5 & 24.7 & 28.4 & 23.4 \\
    Mixtral-8x7B-Instruct-v0.1 \emph{(8-shot)}  & 32.9 & 27.2 & 30.4 & 25.4 \\
    Mixtral-8x7B-Instruct-v0.1 \emph{(16-shot)} & 34.1 & 28.1 & 31.4 & 25.8 \\
    Mixtral-8x7B-Instruct-v0.1 \emph{(32-shot)} & 35.1 & 29.2 & 32.0 & 26.5 \\
    \midrule
    gpt-3.5-turbo-1106 \emph{(2-shot)}          & 33.2 & 25.9 & 30.5 & 23.8 \\
    gpt-3.5-turbo-1106 \emph{(8-shot)}          & 34.9 & 26.9 & 31.8 & 24.7 \\
    \bottomrule

\end{tabular}}
\caption{Average results over all datasets for event argument extraction (EAE) on sampled 250 documents.}
\label{tab:llm_eae}
\vspace{-1em}
\end{table}

\subsection{Analysis}

We also manually examine the cases where LLMs make mistakes. 
The major errors of LLMs can be categorized into the following three cases, suggesting that there is still room for improving LLMs' performance.

\mypar{Overly aggressive predictions.}
We observed that LLMs struggle to accurately capture the concept of certain event types solely from in-context examples, leading to a tendency to generate many false positives.
For instance, considering the following input:
\\[6pt]
\noindent\fbox{\begin{minipage}{0.98\columnwidth}
\textit{Alleged ties to Gulen-In a statement to the United Nations on May 15, the legal Christian advocacy group, American Center for Law and Justice (ACLJ), said Brunson was told that he was being detained as a "national security risk”.}
\end{minipage}}
\\[8pt]
LLMs would predict \emph{detained} as the trigger word for several event types, \emph{Conflict-Attack}, \emph{Life-Die}, \emph{Movement-Transport}, and \emph{Justice-Arrest-Jail}, while the correct event type is only \emph{Justice-Arrest-Jail}. 
This reveals that LLMs might rely heavily on the format of the in-context examples to generate output, rather than fully understanding the semantics of the event types.

\mypar{Imprecise span boundaries.}
We find that another key challenge of generation-based models is to predict accurate offsets. 
For example, considering the following input:
\\[6pt]
\noindent\fbox{\begin{minipage}{0.98\columnwidth}
\textit{In 1988 , Spain supplied Iran with 200,000 respirators.}
\end{minipage}}
\\[8pt]
LLMs would identify \emph{respirators} as the argument of role \emph{Theme}, while the ground truth argument is \emph{200,000 respirators}.

\mypar{Hallucination or paraphrasing.}
We also notice that LLMs may generate spans that are not present in the input text. 
Most of the time, this can be detected by a post-processing script to filter out invalid predictions. 
However, in some cases, LLMs generate reasonable answers but in different textual formats, such as predicting \emph{Los Angeles} when the ground truth is \emph{LA}. 
The current evaluation pipeline would count this as an error.

%% file: 06-future.tex
\section{Future Challenges and Opportunities}
\label{sec:future}
In this section, we discuss the role of event extraction in the current NLP era, as well as some challenges and insights derived from \mytool{}. 

\mypar{How should we position event extraction in the era of LLMs?}
Based on the findings in Section~\ref{sec:llm}, LLMs struggle with extracting and comprehending complicated structured semantic concepts.
This indicates the need for a dedicated system with specialized design to effectively recognize and extract abstract concepts and relations from texts.
We believe that a good event extractor, capable of identifying a wide range of events, could serve as a tool that provides grounded structured information about texts for LLMs. 
Accordingly, LLMs can flexibly decide whether they require this information for the following reasoning steps or inference process.
To achieve this goal, we expect event extractors to be universal, efficient, and accurate, which introduces the following research challenges.

\mypar{Broader event coverage and generalizablity.}
We anticipate that a strong event extractor can recognize a wide range of events  and even identify new event concepts that may not have appeared during training.
This requires two efforts: 
(1) \textit{Expanding domain coverage in datasets}.
Most existing event extraction datasets suffer from a restricted coverage of event types. 
For instance, all the datasets incorporated by \mytool{} have no more than 200 event types, which is significantly below the amount of human concepts encountered in daily life.
Although some recent studies have attempted to tackle this issue~\cite{Li23glen}, their data often contains label noise and lacks detailed role annotations.
We believe that efficiently collecting or synthesizing high-quality data that covers a wild range of events is crucial for enhancing the emerging ability to generalize event recognition.
(2) \textit{Better model design for generalization}.
Most existing event extraction models focus on in-domain performance. 
Therefore, their design can fail when encountering novel events.
While exploring prompting in LLMs shows promise, as discussed in Section~\ref{sec:llm}, the results remain unsatisfactory.
Some recent works \cite{Lu22multi,Ping23uniex} explore learning a unified model across multiple information extraction tasks for improved generalization, but their integration is constrained by limited domains. 
We expect that \mytool{} can serve as a starting point for aggregating diverse datasets and training more robust unified models.

\mypar{Enhanced model efficiency.}
Inference time can pose a bottleneck for effective event extraction, especially when the number of event (role) types increases.
For instance, well-performing methods in \mytool{} (e.g., TagPrime and PAIE) require enumerating all the event (role) types, resulting in multiple times of model inference, which significantly slows down as more events (roles) are considered.
Similar challenges arise with LLMs, as we have to prompt them per event.
Therefore, there is a critical necessity for model designs that not only prioritize performance but also optimize efficiency.

%% file: 10-conclusion.tex
\section{Conclusion}

In this work, we identify and discuss several evaluation issues for event extraction, including inconsistent comparisons, insufficiency, and low reproducibility.
To address these challenges, we propose \mytool{}, a consistent, sufficient, and reproducible benchmark for event extraction.
We also study and benchmark the capability of five large language models in event extraction.
Additionally, we discuss the role of event extraction in the current NLP era, as well as challenges and insights derived from \mytool{}.
We expect \mytool{} and our reevaluation results will serve as a reliable benchmark for research in event extraction.

%% file: 11-limitation.tex
\section*{Acknowledgements}
We thank the anonymous reviewers for their constructive suggestions. 
We also thank UIUC BLENDER Lab, UCLA-NLP group, and UCLA PLUS Lab for the valuable discussions and
comments.
This research is based upon work supported by U.S.~DARPA KAIROS Program No.~FA8750-19-2-1004. 
The views and conclusions contained herein are those of the authors and should not be interpreted as necessarily representing the official policies, either expressed or implied, of DARPA, or the U.S.~Government.
The U.S.~Government is authorized to reproduce and distribute reprints for governmental purposes notwithstanding any copyright annotation therein.

\section*{Limitations}

In this work, we make efforts to incorporate as many event extraction datasets as possible.
However, for some datasets, it is hard for us to obtain the raw files.
Moreover, there is a possibility that we may overlook some datasets.
Similarly, we aim to include a broad range of event extraction approaches, but we acknowledge that it is not feasible to cover all works in the field.
We do our best to consider representative methods that published in recent years.
Additionally, for works without released codebases, we make efforts to reimplement their proposed methods based on the descriptions in the original papers.
There can be discrepancies between our implementation and theirs due to differences in packages and undisclosed techniques.
We will continue to maintain our proposed library and welcome contributions and updates from the community.

%% file: 90-datasets.tex
\section{Details of Dataset Preprocessing}
\label{app:data}

We describe the detailed preprocessing steps for each dataset in the following.
Table~\ref{tab:split1} and~\ref{tab:split2} lists the statistics of each dataset.

\paragraph{ACE05-en \cite{Doddington04ace}.}
We download the ACE05 dataset from LDC\footnote{\url{https://catalog.ldc.upenn.edu/LDC2006T06}} and consider the data in English. The original text in ACE05 dataset is document-based. We follow most prior usage of the dataset \cite{Lin20oneie, Wadden19dygiepp} to split each document into sentences and making it a sentence-level benchmark on event extraction. We use Stanza \cite{Qi20stanza} to perform sentence splitting and discard any label (entity mention, relation mention, event arguments, etc.) where its span is not within a single sentence. Similar to prior works \cite{Lin20oneie, Wadden19dygiepp}, we consider using head span to represent entity mentions and only include event arguments that are entities (i.e., remove time and values in the ACE05 annotation). The original annotation of the dataset is character-level. However, to make the dataset consistent with others, we perform tokenization through Stanza and map the character-level annotation into token-level.
We split the train, dev, and test sets based on documents with the ratio 80\%, 10\%, and 10\%.

\paragraph{RichERE \cite{Song15ere}.}
Considering the unavailability of the RichERE dataset used in prior works \cite{Lin20oneie, Hsu22degree}, we download the latest RichERE dataset from LDC\footnote{\url{https://catalog.ldc.upenn.edu/LDC2023T04}} and only consider the 288 documents labeled with Rich ERE annotations. Similar to the pre-processing step in ACE05-en, we use Stanza \cite{Qi20stanza} to perform sentence splitting and making it a sentence-level benchmark. Following the strategy in \cite{Lin20oneie}, we use head span to represent entity mentions and only consider named entities, weapons and vehicles as event argument candidates. Again, the original annotation of the dataset is character-level, and we perform tokenization through Stanza and map the annotation into token-level, forming the final RichERE dataset we use. 
We split the train, dev, and test sets based on documents with the ratio 80\%, 10\%, and 10\%.

\paragraph{MLEE \cite{Pyysalo12mlee}.}
The original MLEE dataset is document-level.\footnote{\url{https://www.nactem.ac.uk/MLEE/}}
We use Stanza \cite{Qi20stanza} to do the sentence tokenization and the word tokenization.
For the purpose of evaluating most baselines, we divide the documents into several segment-level instances with a sub-token window size being 480 based on the RoBERTa-large tokenizer \cite{Liu19roberta}.
We split the train, dev, and test sets based on documents with the ratio 70\%, 15\%, and 15\%.

\paragraph{Genia2011 \cite{Kim11genia2011}.}
The original Genia2011 dataset is document-level.\footnote{\url{https://bionlp-st.dbcls.jp/GE/2011/downloads/}}
We use Stanza \cite{Qi20stanza} to do the sentence tokenization and the word tokenization.
For the purpose of evaluating most baselines, we divide the documents into several segment-level instances with a sub-token window size being 480 based on the RoBERTa-large tokenizer \cite{Liu19roberta}.
We split the train, dev, and test sets based on documents with the ratio 60\%, 20\%, and 20\%.

\paragraph{Genia2013 \cite{Kim13genia2013}.}
The original Genia2013 dataset is document-level.\footnote{\url{https://2013.bionlp-st.org/tasks/}}
We use Stanza \cite{Qi20stanza} to do the sentence tokenization and the word tokenization.
For the purpose of evaluating most baselines, we divide the documents into several segment-level instances with a sub-token window size being 480 based on the RoBERTa-large tokenizer \cite{Liu19roberta}.
We split the train, dev, and test sets based on documents with the ratio 60\%, 20\%, and 20\%.

\paragraph{M$^2$E$^2$ \cite{Li20m2e2}.}
The M$^2$E$^2$ dataset contains event argument annotations from both texts and images.\footnote{\url{https://blender.cs.illinois.edu/software/m2e2}} 
We consider only the text annotations in our benchmark.
We directly use the tokenized words without any modifications.
We merge the original train, dev, and test sets, and split them into the new train, dev, and test sets based on documents with the ratio 70\%, 15\%, and 15\%.

\paragraph{CASIE \cite{Satyapanich20casie}.}
The original CASIE dataset is document-level.\footnote{\url{https://github.com/Ebiquity/CASIE}}
We use Stanza \cite{Qi20stanza} to do the sentence tokenization and the word tokenization.
For the purpose of evaluating most baselines, we divide the documents into several segment-level instances with a sub-token window size being 480 based on the RoBERTa-large tokenizer \cite{Liu19roberta}.
We split the train, dev, and test sets based on documents with the ratio 70\%, 15\%, and 15\%.

\paragraph{PHEE \cite{Sun22phee}.}
We download the PHEE dataset from the official webpage.\footnote{\url{https://github.com/ZhaoyueSun/PHEE}}
We directly use the tokenized words without any modifications.
We merge the original train, dev, and test sets, and split them into the new train, dev, and test sets based on documents with the ratio 60\%, 20\%, and 20\%.

\paragraph{MAVEN \cite{Wang20maven}.}
We consider the sentence-level annotations from the original data.\footnote{\url{https://github.com/THU-KEG/MAVEN-dataset}}
We directly use the tokenized words without any modifications.
Because the labels of the original test set are not publicly accessible, we merge the original train and dev sets and split it into new train, dev, and test sets by documents with the ratio 70\%, 15\%, and 15\%.

\paragraph{MEE-en \cite{Veyseh22mee}.}
We download the MEE dataset\footnote{\url{http://nlp.uoregon.edu/download/MEE/MEE.zip}} and consider the English annotations.
We use the annotations for event detection  only because we observe that the quality of the annotations for event argument extraction is not good and many important arguments are actually missing.
We directly use the tokenized words without any modifications.
We merge the original train, dev, and test sets, and split them into the new train, dev, and test sets based on documents with the ratio 80\%, 10\%, and 10\%.

\paragraph{FewEvent \cite{Deng20fewevent}.}
We download the FewEvent dataset from the official webpage.\footnote{\url{https://github.com/231sm/Low_Resource_KBP}}
Notice that we consider FewEvent as a normal supervised event detection dataset.
We use Stanza \cite{Qi20stanza} to do the word tokenization.
For the purpose of evaluating most baselines, we discard the instances with the length longer than 300.
We split the train, dev, and test sets based on documents with the ratio 60\%, 20\%, and 20\%.

\paragraph{SPEED \cite{Parekh2024speed}.}
We download the SPEED dataset from the official webpage.\footnote{\url{https://github.com/PlusLabNLP/SPEED}}
Notice that we consider only the COVID-related examples.
We split the train, dev, and test sets based on documents with the ratio 60\%, 20\%, and 20\%.

\paragraph{RAMS \cite{Ebner20rams}.}
We use the latest version of the RAMS dataset.\footnote{\url{https://nlp.jhu.edu/rams/RAMS_1.0c.tar.gz}}
We directly use the tokenized words without any modifications.
For the purpose of evaluating most baselines, we discard the instances with the sub-token length larger than 500 based on the RoBERTa-large tokenizer \cite{Liu19roberta}.
We merge the original train, dev, and test sets, and split them into the new train, dev, and test sets based on documents with the ratio 80\%, 10\%, and 10\%.

\paragraph{WikiEvents \cite{Li21wikievents}.}
We download the WikiEvents dataset from the official webpage.\footnote{\url{s3://gen-arg-data/wikievents/}}
We directly use the tokenized words without any modifications.
For the purpose of evaluating most baselines, we divide the documents into several segment-level instances with a sub-token window size being 480 based on the RoBERTa-large tokenizer \cite{Liu19roberta}.
We split the train, dev, and test sets based on documents with the ratio 80\%, 10\%, and 10\%.

\paragraph{MUC-4 \cite{muc1992muc4}.}
We use the preprocessed data from the GRIT repository.\footnote{\url{https://github.com/xinyadu/grit_doc_event_entity/}}
We use Stanza \cite{Qi20stanza} to do the sentence tokenization and the word tokenization.
For the purpose of evaluating most baselines, we divide the documents into several segment-level instances with a sub-token window size being 480 based on the RoBERTa-large tokenizer \cite{Liu19roberta}.
We split the train, dev, and test sets based on documents with the ratio 60\%, 20\%, and 20\%.

\paragraph{GENEVA \cite{parekh2023geneva}.}
We download the GENEVA dataset from the officail webpage.\footnote{\url{https://github.com/PlusLabNLP/GENEVA}}
We directly use the tokenized words without any modifications.
We split the train, dev, and test sets based on documents with the ratio 70\%, 15\%, and 15\%.

\begin{table*}[t!]
\centering
\setlength{\tabcolsep}{2.5pt}
\resizebox{.9\textwidth}{!}{
\begin{tabular}{c|c|c|cccccc|cccccc|cccccc}
    \toprule
    \multirow{2}{*}{Dataset} & \multirow{2}{*}{Task} & \multirow{2}{*}{Split}
    & \multicolumn{6}{c|}{\textbf{Train}} & \multicolumn{6}{c|}{\textbf{Dev}} & \multicolumn{6}{c}{\textbf{Test}}
    \\
    & & & \#Docs & \#Inst & \#ET & \#Evt & \#RT & \#Arg 
    & \#Docs & \#Inst & \#ET & \#Evt & \#RT & \#Arg 
    & \#Docs & \#Inst & \#ET & \#Evt & \#RT & \#Arg \\
    \midrule
    \multirow{5}{*}{ACE05-en} & \multirow{5}{*}{\makecell{E2E \\ ED \\ EAE}} & 1
    &  481  &   16531    &     33       &      4309        &      22     &    6503    
    &   59  &    1870    &     30       &       476        &      22     &     766    
    &   59  &    2519    &     30       &       563        &      22     &     828    \\
    & & 2
    &  481  &   17423    &     33       &      4348        &      22     &    6544    
    &   59  &    1880    &     29       &       555        &      22     &     894    
    &   59  &    1617    &     30       &       445        &      22     &     659    \\
    & & 3
    &  481  &   17285    &     33       &      4331        &      22     &    6484    
    &   59  &    2123    &     30       &       515        &      22     &     799    
    &   59  &    1512    &     30       &       502        &      22     &     814    \\
    & & 4
    &  481  &   16842    &     33       &      4437        &      22     &    6711    
    &   59  &    1979    &     30       &       460        &      22     &     728    
    &   59  &    2099    &     29       &       451        &      22     &     658    \\
    & & 5
    &  481  &   16355    &     33       &      4198        &      22     &    6392    
    &   59  &    1933    &     30       &       509        &      22     &     772    
    &   59  &    2632    &     31       &       641        &      22     &     933    \\
    \midrule
    \multirow{5}{*}{RichERE} & \multirow{5}{*}{\makecell{E2E \\ ED \\ EAE}} & 1 
    &  232  &    9198    &     38       &      4549        &      21     &    6581    
    &   28  &     876    &     35       &       488        &      21     &     737    
    &   28  &    1167    &     34       &       672        &      21     &     936    \\
    & & 2
    &  232  &    8886    &     38       &      4444        &      21     &    6520    
    &   28  &    1299    &     36       &       688        &      21     &     978    
    &   28  &    1056    &     37       &       577        &      21     &     756    \\
    & & 3
    &  232  &    9094    &     38       &      4490        &      21     &    6517    
    &   28  &    1081    &     36       &       678        &      21     &     942    
    &   28  &    1066    &     35       &       541        &      21     &     795    \\
    & & 4
    &  232  &    9105    &     38       &      4541        &      21     &    6647    
    &   28  &     973    &     34       &       571        &      21     &     804    
    &   28  &    1163    &     37       &       597        &      21     &     803    \\
    & & 5
    &  232  &    9169    &     38       &      4682        &      21     &    6756    
    &   28  &    1135    &     34       &       487        &      21     &     692    
    &   28  &     937    &     35       &       540        &      21     &     806    \\
    \midrule
    \multirow{5}{*}{MLEE} & \multirow{5}{*}{\makecell{E2E \\ ED \\ EAE}} & 1
    &  184  &     199    &      29      &       4705       &      14     &    4237    
    &   39  &      45    &      21      &       1003       &       9     &     895    
    &   39  &      42    &      21      &        867       &      12     &     826    \\
    & & 2
    &  184  &     202    &      29      &       4733       &      14     &    4258    
    &   39  &      42    &      19      &        898       &      10     &     854    
    &   39  &      42    &      21      &        944       &      11     &     846    \\
    & & 3
    &  184  &     200    &      29      &       4627       &      14     &    4165    
    &   39  &      42    &      20      &       1029       &      10     &     944    
    &   39  &      44    &      20      &        919       &      10     &     849    \\
    & & 4
    &  184  &     203    &      29      &       4629       &      14     &    4236    
    &   39  &      40    &      20      &        980       &      11     &     872    
    &   39  &      43    &      20      &        966       &      11     &     850    \\
    & & 5
    &  184  &     201    &      29      &       4653       &      14     &    4200    
    &   39  &      42    &      21      &        887       &      11     &     843    
    &   39  &      43    &      20      &       1035       &      11     &     915    \\
    \midrule
    \multirow{5}{*}{Genia2011} & \multirow{5}{*}{\makecell{E2E \\ ED \\ EAE}} & 1
    &  576  &     773    &       9      &       7396       &      10     &    6495
    &  192  &     348    &       9      &       3773       &       9     &    3352
    &  192  &     254    &       9      &       2368       &       8     &    2018   \\
    & & 2
    &  576  &     843    &       9      &       8455       &      10     &    7397
    &  192  &     266    &       9      &       2713       &       9     &    2358
    &  192  &     266    &       9      &       2369       &       9     &    2110   \\
    & & 3
    &  576  &     901    &       9      &       8638       &      10     &    7687
    &  192  &     233    &       9      &       2042       &       8     &    1743
    &  192  &     241    &       9      &       2857       &       9     &    2435   \\
    & & 4
    &  576  &     808    &       9      &       7836       &      10     &    7037
    &  192  &     277    &       9      &       2842       &       9     &    2319
    &  192  &     290    &       9      &       2859       &       9     &    2509   \\
    & & 5
    &  576  &     853    &       9      &       8460       &      10     &    7464
    &  192  &     240    &       9      &       2368       &       9     &    2061
    &  192  &     282    &       9      &       2709       &       9     &    2340   \\
    \midrule
    \multirow{5}{*}{Genia2013} & \multirow{5}{*}{\makecell{E2E \\ ED \\ EAE}} & 1
    &   12  &     420    &      13      &       4077       &       7     &    3921    
    &   4   &     105    &      10      &        950       &       7     &     858    
    &   4   &     139    &      11      &        974       &       7     &     881    \\
    & & 2
    &   12  &     388    &      13      &       3578       &       7     &    3561    
    &   4   &     128    &      11      &       1284       &       6     &    1134    
    &   4   &     148    &      10      &       1149       &       6     &     965    \\
    & & 3
    &   12  &     381    &      13      &       3816       &       7     &    3674    
    &   4   &     143    &      10      &       1174       &       7     &    1079    
    &   4   &     140    &      11      &       1011       &       6     &     907    \\
    & & 4
    &   12  &     441    &      13      &       3971       &       7     &    3993    
    &   4   &     111    &       9      &        785       &       7     &     616    
    &   4   &     112    &      11      &       1245       &       6     &    1051    \\
    & & 5
    &   12  &     427    &      13      &       4225       &       7     &    4112    
    &   4   &     120    &      10      &        809       &       6     &     717    
    &   4   &     117    &      10      &        967       &       7     &     831    \\
    \midrule
    \multirow{5}{*}{M$^2$E$^2$} & \multirow{5}{*}{\makecell{E2E \\ ED \\ EAE}} & 1
    &  4211 &    4211    &      8       &       748        &      15     &    1120    
    &   901 &     901    &      8       &       183        &      15     &     280    
    &   901 &     901    &      8       &       174        &      15     &     259    \\
    & & 2
    &  4211 &    4211    &      8       &       794        &      15     &    1171    
    &   901 &     901    &      8       &       148        &      14     &     232    
    &   901 &     901    &      8       &       163        &      15     &     256    \\
    & & 3
    &  4211 &    4211    &      8       &       760        &      15     &    1138    
    &   901 &     901    &      8       &       160        &      15     &     252    
    &   901 &     901    &      8       &       185        &      15     &     269    \\
    & & 4
    &  4211 &    4211    &      8       &       770        &      15     &    1137    
    &   901 &     901    &      8       &       178        &      15     &     276    
    &   901 &     901    &      8       &       157        &      15     &     246    \\
    & & 5
    &  4211 &    4211    &      8       &       747        &      15     &    1122    
    &   901 &     901    &      8       &       164        &      14     &     258    
    &   901 &     901    &      8       &       194        &      15     &     279    \\
    \midrule
    \multirow{5}{*}{CASIE} & \multirow{5}{*}{\makecell{E2E \\ ED \\ EAE}} & 1
    &  701  &    1047    &      5       &       5980       &      26     &    15869   
    &  149  &    218     &      5       &       1221       &      26     &    3175    
    &  149  &    218     &      5       &       1268       &      26     &    3531    \\
    & & 2
    &  701  &    1046    &      5       &       6010       &      26     &    15986   
    &  149  &    223     &      5       &       1294       &      26     &    3492    
    &  149  &    214     &      5       &       1165       &      26     &    3097    \\
    & & 3
    &  701  &    1044    &      5       &       6009       &      26     &    16090   
    &  149  &    210     &      5       &       1286       &      26     &    3344    
    &  149  &    229     &      5       &       1174       &      26     &    3141    \\
    & & 4
    &  701  &    1040    &      5       &       6034       &      26     &    15962   
    &  149  &    229     &      5       &       1172       &      26     &    3211    
    &  149  &    214     &      5       &       1263       &      26     &    3402    \\
    & & 5
    &  701  &    1043    &      5       &       5831       &      26     &    15544   
    &  149  &    218     &      5       &       1288       &      26     &    3369    
    &  149  &    222     &      5       &       1350       &      26     &    3662    \\
    \midrule
    \multirow{5}{*}{PHEE} & \multirow{5}{*}{\makecell{E2E \\ ED \\ EAE}} & 1
    & 2897  &    2897    &       2      &       3003       &      16     &    15482   
    &  965  &     965    &       2      &       1011       &      16     &     5123   
    &  965  &     965    &       2      &       1005       &      16     &     5155   \\
    & & 2
    & 2897  &    2897    &       2      &       3014       &      16     &    15576   
    &  965  &     965    &       2      &       1002       &      16     &     5090   
    &  965  &     965    &       2      &       1003       &      16     &     5094   \\
    & & 3
    & 2897  &    2897    &       2      &       3009       &      16     &    15230   
    &  965  &     965    &       2      &       1001       &      16     &     5200   
    &  965  &     965    &       2      &       1009       &      16     &     5330   \\
    & & 4
    & 2897  &    2897    &       2      &       3020       &      16     &    15496   
    &  965  &     965    &       2      &        996       &      16     &     5124   
    &  965  &     965    &       2      &       1003       &      16     &     5140   \\
    & & 5
    & 2897  &    2897    &       2      &       3011       &      16     &    15498   
    &  965  &     965    &       2      &       1000       &      16     &     5049   
    &  965  &     965    &       2      &       1008       &      16     &     5213   \\
    \bottomrule
\end{tabular}}
\caption{Detailed statistics of each data split for E2E datasets. \emph{\#Docs, \#Inst, \#ET, \#EvT, \#RT, and \#Arg represent the number of documents, instances, event types, events, roles, and arguments, respectively.}}
\label{tab:split1}
\vspace{-1em}
\end{table*}

\begin{table*}[t!]
\centering
\setlength{\tabcolsep}{2.5pt}
\resizebox{.9\textwidth}{!}{
\begin{tabular}{c|c|c|cccccc|cccccc|cccccc}
    \toprule
    \multirow{2}{*}{Dataset} & \multirow{2}{*}{Task} & \multirow{2}{*}{Split}
    & \multicolumn{6}{c|}{\textbf{Train}} & \multicolumn{6}{c|}{\textbf{Dev}} & \multicolumn{6}{c}{\textbf{Test}}
    \\
    & & & \#Docs & \#Inst & \#ET & \#Evt & \#RT & \#Arg 
    & \#Docs & \#Inst & \#ET & \#Evt & \#RT & \#Arg 
    & \#Docs & \#Inst & \#ET & \#Evt & \#RT & \#Arg \\
    \midrule
    \multirow{5}{*}{MAVEN} & \multirow{5}{*}{ED} & 1
    &  2537 &    28734   &      168     &       69069   & -- & --   
    &  543  &    5814    &      167     &       13638   & -- & --   
    &  543  &    5925    &      168     &       14190   & -- & --   \\
    & & 2
    &  2537 &    28341   &      168     &       68162   & -- & --   
    &  543  &    5982    &      167     &       14233   & -- & --   
    &  543  &    6150    &      168     &       14502   & -- & --   \\
    & & 3
    &  2537 &    28348   &      168     &       67832   & -- & --   
    &  543  &    6049    &      167     &       14185   & -- & --   
    &  543  &    6076    &      168     &       14880   & -- & --   \\
    & & 4
    &  2537 &    28172   &      168     &       67450   & -- & --   
    &  543  &    6190    &      167     &       14637   & -- & --   
    &  543  &    6111    &      167     &       14810   & -- & --   \\
    & & 5
    &  2537 &    28261   &      168     &       67826   & -- & --   
    &  543  &    6190    &      167     &       14493   & -- & --   
    &  543  &    6022    &      168     &       14578   & -- & --   \\
    \midrule
    \multirow{5}{*}{MEE-en} & \multirow{5}{*}{ED} & 1
    & 10400 &    10400   &      16      &       13748      &   --   &  --   
    &  1300 &    1300    &      16      &       1764       &   --   &  --   
    &  1300 &    1300    &      16      &       1745       &   --   &  --   \\
    & & 2
    & 10400 &    10400   &      16      &       13801      &   --   &  --   
    &  1300 &    1300    &      16      &       1731       &   --   &  --   
    &  1300 &    1300    &      16      &       1725       &   --   &  --   \\
    & & 3
    & 10400 &    10400   &      16      &       13847      &   --   &  --   
    &  1300 &    1300    &      16      &       1722       &   --   &  --   
    &  1300 &    1300    &      16      &       1688       &   --   &  --   \\
    & & 4
    & 10400 &    10400   &      16      &       13855      &   --   &  --   
    &  1300 &    1300    &      16      &       1701       &   --   &  --   
    &  1300 &    1300    &      16      &       1701       &   --   &  --   \\
    & & 5
    & 10400 &    10400   &      16      &       13802      &   --   &  --   
    &  1300 &    1300    &      16      &       1734       &   --   &  --   
    &  1300 &    1300    &      16      &       1721       &   --   &  --   \\
    \midrule
    \multirow{5}{*}{FewEvent} & \multirow{5}{*}{ED} & 1
    &  7579 &    7579    &      100     &       7579       & -- & --
    &  2513 &    2513    &      98      &       2513       & -- & --
    &  2541 &    2541    &      99      &       2541       & -- & -- \\
    & & 2
    &  7579 &    7579    &      100     &       7579       & -- & --
    &  2513 &    2513    &      98      &       2513       & -- & --
    &  2541 &    2541    &      99      &       2541       & -- & -- \\
    & & 3
    &  7579 &    7579    &      100     &       7579       & -- & --
    &  2513 &    2513    &      98      &       2513       & -- & --
    &  2541 &    2541    &      99      &       2541       & -- & -- \\
    & & 4
    &  7579 &    7579    &      100     &       7579       & -- & --
    &  2513 &    2513    &      98      &       2513       & -- & --
    &  2541 &    2541    &      99      &       2541       & -- & -- \\
    & & 5
    &  7579 &    7579    &      100     &       7579       & -- & --
    &  2513 &    2513    &      98      &       2513       & -- & --
    &  2541 &    2541    &      99      &       2541       & -- & -- \\
    \midrule
    \multirow{5}{*}{SPEED} & \multirow{5}{*}{ED} & 1
    & 1185  &    1185    &      7       &       1334       & -- & --
    & 395   &    395     &      7       &       415        & -- & --
    & 395   &    395     &      7       &       458        & -- & -- \\
    & & 2
    & 1185  &    1185    &      7       &       1361       & -- & --
    & 395   &    395     &      7       &       432        & -- & --
    & 395   &    395     &      7       &       424        & -- & -- \\
    & & 3
    & 1185  &    1185    &      7       &       1336       & -- & --
    & 395   &    395     &      7       &       449        & -- & --
    & 395   &    395     &      7       &       432        & -- & -- \\
    & & 4
    & 1185  &    1185    &      7       &       1328       & -- & --
    & 395   &    395     &      7       &       460        & -- & --
    & 395   &    395     &      7       &       429        & -- & -- \\
    & & 5
    & 1185  &    1185    &      7       &       1340       & -- & --
    & 395   &    395     &      7       &       446        & -- & --
    & 395   &    395     &      7       &       431        & -- & -- \\
    \midrule
    \multirow{5}{*}{RAMS} & \multirow{5}{*}{EAE} & 1
    &  7827 &    7827    &     139      &       7287       &      65     &    16951   
    &   910 &    910     &     136      &       910        &      64     &    2132    
    &   910 &    910     &     135      &       910        &      63     &    2123    \\
    & & 2
    &  7827 &    7827    &     139      &       7287       &      65     &    16946   
    &   910 &    910     &     135      &       910        &      65     &    2113    
    &   910 &    910     &     137      &       910        &      65     &    2147    \\
    & & 3
    &  7827 &    7827    &     139      &       7287       &      65     &    16937   
    &   910 &    910     &     135      &       910        &      64     &    2168    
    &   910 &    910     &     135      &       910        &      64     &    2101    \\
    & & 4
    &  7827 &    7827    &     139      &       7287       &      65     &    17014   
    &   910 &    910     &     136      &       910        &      62     &    2093    
    &   910 &    910     &     137      &       910        &      63     &    2099    \\
    & & 5
    &  7827 &    7827    &     139      &       7287       &      65     &    17003   
    &   910 &    910     &     135      &       910        &      63     &    2130    
    &   910 &    910     &     137      &       910        &      65     &    2073    \\
    \midrule
    \multirow{5}{*}{WikiEvents} & \multirow{5}{*}{EAE} & 1
    &  197  &    450     &      50      &       3131       &      57     &    4393    
    &   24  &     53     &      39      &       422        &      43     &     592    
    &   24  &     62     &      38      &       379        &      46     &     516    \\
    & & 2
    &  197  &    439     &      50      &       2990       &      57     &    4234    
    &   24  &     57     &      39      &       405        &      42     &     571    
    &   24  &     69     &      37      &       537        &      38     &     696    \\
    & & 3
    &  197  &    435     &      50      &       3014       &      56     &    4228    
    &   24  &     78     &      36      &       471        &      43     &     623    
    &   24  &     52     &      37      &       447        &      47     &     650    \\
    & & 4
    &  197  &    454     &      50      &       3143       &      57     &    4391    
    &   24  &     46     &      36      &       431        &      43     &     606    
    &   24  &     65     &      40      &       358        &      47     &     504    \\
    & & 5
    &  197  &    441     &      50      &       3142       &      57     &    4370    
    &   24  &     57     &      38      &       394        &      43     &     562    
    &   24  &     67     &      40      &       396        &      45     &     569    \\
    \midrule
    \multirow{5}{*}{MUC-4} & \multirow{5}{*}{EAE} & 1
    &  1020 &    1407    &      1       &       1407       &      5      &    2974    
    &  340  &    489     &      1       &        489       &      5      &     918    
    &  340  &    464     &      1       &        464       &      5      &     884    \\
    & & 2
    &  1020 &    1408    &      1       &       1408       &      5      &    2990    
    &  340  &    489     &      1       &        489       &      5      &     897    
    &  340  &    463     &      1       &        463       &      5      &     889    \\
    & & 3
    &  1020 &    1419    &      1       &       1419       &      5      &    2912    
    &  340  &    473     &      1       &        473       &      5      &     994    
    &  340  &    468     &      1       &        468       &      5      &     870    \\
    & & 4
    &  1020 &    1425    &      1       &       1425       &      5      &    2889    
    &  340  &    475     &      1       &        475       &      5      &     921    
    &  340  &    460     &      1       &        460       &      5      &     966    \\
    & & 5
    &  1020 &    1427    &      1       &       1427       &      5      &    2928    
    &  340  &    465     &      1       &        465       &      5      &     929    
    &  340  &    468     &      1       &        468       &      5      &     919    \\
    \midrule
    \multirow{5}{*}{GENEVA} & \multirow{5}{*}{EAE} & 1
    &  96   &    2582    &      115     &       5290       &     220     &    8618    
    &  82   &    509     &      115     &       1016       &     159     &    1683    
    &  84   &    593     &      115     &       1199       &     171     &    2013    \\
    & & 2
    &  97   &    2583    &      115     &       5268       &     220     &    8660    
    &  85   &    509     &      114     &       1014       &     158     &    1615    
    &  85   &    592     &      115     &       1223       &     164     &    1994    \\
    & & 3
    &  97   &    2582    &      115     &       5294       &     220     &    8638    
    &  85   &    509     &      115     &       1010       &     156     &    1642    
    &  81   &    593     &      115     &       1201       &     170     &    1989    \\
    & & 4
    &  96   &    2582    &      115     &       5293       &     220     &    8705    
    &  79   &    509     &      115     &       1003       &     164     &    1636    
    &  88   &    593     &      115     &       1209       &     166     &    1928    \\
    & & 5
    &  97   &    2582    &      115     &       5337       &     220     &    8673    
    &  88   &    509     &      115     &       1004       &     161     &    1680    
    &  86   &    593     &      115     &       1164       &     161     &    1916    \\
    \bottomrule
\end{tabular}}
\caption{Detailed statistics of each data split for ED and EAE datasets. \emph{\#Docs, \#Inst, \#ET, \#EvT, \#RT, and \#Arg represent the number of documents, instances, event types, events, roles, and arguments, respectively.}}
\label{tab:split2}
\vspace{-1em}
\end{table*}

%% file: 91-models.tex
\section{Details of Model Implementations}
\label{app:models}

We utilize RoBERTa-large \cite{Liu19roberta} for all the classification-based models and use BART-large \cite{Lewis20bart} for all the generation-based models to have a consistent comparison.

\paragraph{DyGIE++ \cite{Wadden19dygiepp}.}
We re-implement the model based on the original codebase.\footnote{\url{https://github.com/dwadden/dygiepp}}

\paragraph{OneIE \cite{Lin20oneie}.}
We adapt the code from the original codebase.\footnote{\url{https://blender.cs.illinois.edu/software/oneie/}}

\paragraph{AMR-IE \cite{Zhang21amrie}.}
We adapt the code from the original codebase.\footnote{\url{https://github.com/zhangzx-uiuc/AMR-IE}}

\paragraph{EEQA \cite{Du20bertqa}.}
We re-implement the model based on the original codebase.\footnote{\url{https://github.com/xinyadu/eeqa}}
Notice that EEQA requires some human-written queries for making predictions.
For those datasets that EEQA provides queries, we directly use those queries.
For other datasets, we follow the suggestion from the paper and use ``arg'' style queries like ``\{\emph{role\_name}\} in [Trigger]''.

\paragraph{RCEE \cite{Liu20rcee}.}
We re-implement the model based on the description in the original paper.
Notice that RCEE requires a question generator to generate queries for making predictions.
Alternatively, we re-use the queries from EEQA as the generated queries.

\paragraph{Query\&Extract \cite{Wang22queryextract}.}
We adapt the code from the original codebase.\footnote{\url{https://github.com/VT-NLP/Event_Query_Extract}}
We use the event type names as the verbalized string for each event.
Since the origin model supports event argument role labeling rather than event argument extraction, we learn an additional NER sequential labeling model during training and use the predicted entities for event argument role labeling during testing.

\paragraph{TagPrime \cite{Hsu22tagprime}.}
We adapt the code from the original codebase.\footnote{\url{https://github.com/PlusLabNLP/TagPrime}}

\paragraph{PAIE \cite{Ma22paie}.}
We adapt the code from the original codebase.\footnote{\url{https://github.com/mayubo2333/PAIE}}
Notice that PAIE requires some human-written templates for making predictions.
For those datasets that PAIE provides templates, we directly use them.
For other datasets, we create automated templates like ``\{\emph{role\_1\_name}\} [argument\_1] \{\emph{role\_2\_name}\} [argument\_2] ... \{\emph{role\_k\_name}\} [argument\_k] ''.

\paragraph{DEGREE \cite{Hsu22degree}.}
We adapt the code from the original codebase.\footnote{\url{https://github.com/PlusLabNLP/DEGREE}}
Notice that DEGREE requires some human-written templates for making predictions.
For those datasets that DEGREE provides templates, we directly use them.
For other datasets, we re-use the templates generated by PAIE.

\paragraph{BART-Gen \cite{Li21wikievents}.}
We re-implement the model from the original codebase.\footnote{\url{https://github.com/raspberryice/gen-arg}}
We replace the original pure copy mechanism with a copy-generator since we observe this works better.
Notice that BART-Gen requires some human-written templates for making predictions.
For those datasets that BART-Gen provides templates, we directly use them.
For other datasets, we re-use the templates generated by PAIE.

\paragraph{X-Gear \cite{Huang22xgear}.}
We adapt the code from the original codebase.\footnote{\url{https://github.com/PlusLabNLP/X-Gear}}

\paragraph{AMPERE \cite{Hsu23ampere}.}
We adapt the code from the original codebase.\footnote{\url{https://github.com/PlusLabNLP/AMPERE}}
Notice that AMPERE requires some human-written templates for making predictions.
For those datasets that AMPERE provides templates, we directly use them.
For other datasets, we re-use the templates generated by PAIE.

\paragraph{UniST \cite{Huang22unist}.}
We re-implement the model from the original codebase.\footnote{\url{https://github.com/luka-group/unist}}
Since the origin model supports semantic typing only, we learn an additional span recognition model during training and use the predicted trigger spans for trigger span typing during testing.

\paragraph{CEDAR \cite{Li23glen}.}
We re-implement the model from the original codebase.\footnote{\url{https://github.com/ZQS1943/GLEN}}
Notice that in the original paper, they consider \emph{self-labeling} during training as the dataset they consider is noisy.
Our implementation currently ignores the \emph{self-labeling} part.

%% file: 92-results.tex
\section{Detailed Results}
\label{app:results}

Table~\ref{tab:all_e2e}, \ref{tab:all_ed}, \ref{tab:all_eae} demonstrate the detailed reevaluation results for end-to-end event extraction, event detection, and event argument extraction, respectively.

\begin{table*}[t!]
\centering
\setlength{\tabcolsep}{4pt}
\resizebox{.95\textwidth}{!}{
\begin{tabular}{l|cccccc|cccccc|cccccc}
    \toprule
    \multirow{2}{*}{Model} & \multicolumn{6}{c|}{\textbf{ACE05}} & \multicolumn{6}{c|}{\textbf{RichERE}} & \multicolumn{6}{c}{\textbf{MLEE}} \\
    & TI & TC & AI & AC & AI+ & AC+ & TI & TC & AI & AC & AI+ & AC+ & TI & TC & AI & AC & AI+ & AC+ \\
    \midrule
    DyGIE++
    & \cellcolor{red!16}74.7 & \cellcolor{red!24}71.3 & \cellcolor{red!08}59.1 & \cellcolor{red!00}56.0 & \cellcolor{red!08}54.5 & \cellcolor{red!08}51.8 
    & \cellcolor{red!08}69.7 & \cellcolor{red!00}59.8 & \cellcolor{red!00}47.1 & \cellcolor{red!00}42.0 & \cellcolor{red!00}43.1 & \cellcolor{red!00}38.3 
    & \cellcolor{red!16}82.6 & \cellcolor{red!08}78.2 & \cellcolor{red!16}60.4 & \cellcolor{red!16}57.8 & \cellcolor{red!16}56.6 & \cellcolor{red!16}54.4 
    \\
    OneIE
    & \cellcolor{red!24}75.0 & \cellcolor{red!08}71.1 & \cellcolor{red!16}62.4 & \cellcolor{red!16}59.9 & \cellcolor{red!16}56.9 & \cellcolor{red!24}54.7 
    & \cellcolor{red!24}71.0 & \cellcolor{red!16}62.5 & \cellcolor{red!16}53.9 & \cellcolor{red!16}50.0 & \cellcolor{red!16}48.4 & \cellcolor{red!16}45.2 
    & \cellcolor{red!24}82.7 & \cellcolor{red!16}78.5 & \cellcolor{red!00}28.7 & \cellcolor{red!00}26.9 & \cellcolor{red!00}13.6 & \cellcolor{red!00}13.1 
    \\
    AMR-IE
    & \cellcolor{red!08}74.6 & \cellcolor{red!16}71.1 & \cellcolor{red!24}63.1 & \cellcolor{red!24}60.6 & \cellcolor{red!24}56.9 & \cellcolor{red!08}54.6 
    & \cellcolor{red!16}70.5 & \cellcolor{red!08}62.3 & \cellcolor{red!08}53.7 & \cellcolor{red!08}49.5 & \cellcolor{red!08}48.1 & \cellcolor{red!08}44.7 
    & \cellcolor{red!08}82.4 & \cellcolor{red!08}78.2 & \cellcolor{red!00}21.3 & \cellcolor{red!00}15.2 & \cellcolor{red!00}6.0 & \cellcolor{red!00}4.7 
    \\
    EEQA
    & \cellcolor{red!00}73.8 & \cellcolor{red!00}70.0 & \cellcolor{red!00}57.0 & \cellcolor{red!00}55.3 & \cellcolor{red!00}51.9 & \cellcolor{red!00}50.4 
    & \cellcolor{red!00}69.3 & \cellcolor{red!00}60.2 & \cellcolor{red!00}49.2 & \cellcolor{red!00}45.8 & \cellcolor{red!00}44.7 & \cellcolor{red!00}41.9 
    & \cellcolor{red!00}81.4 & \cellcolor{red!00}76.9 & \cellcolor{red!08}52.9 & \cellcolor{red!08}51.1 & \cellcolor{red!08}39.0 & \cellcolor{red!08}38.1 
    \\
    RCEE
    & \cellcolor{red!08}74.0 & \cellcolor{red!08}70.5 & \cellcolor{red!00}57.2 & \cellcolor{red!00}55.5 & \cellcolor{red!00}52.5 & \cellcolor{red!00}51.0 
    & \cellcolor{red!00}68.6 & \cellcolor{red!00}60.0 & \cellcolor{red!00}49.8 & \cellcolor{red!00}46.2 & \cellcolor{red!00}45.1 & \cellcolor{red!00}42.1 
    & \cellcolor{red!00}81.3 & \cellcolor{red!00}77.2 & \cellcolor{red!08}52.0 & \cellcolor{red!00}49.3 & \cellcolor{red!00}36.9 & \cellcolor{red!00}35.4 
    \\
    Query\&Extract
    & \cellcolor{red!00}68.6 & \cellcolor{red!00}65.1 & \cellcolor{red!00}57.4 & \cellcolor{red!00}55.0 & \cellcolor{red!00}51.2 & \cellcolor{red!00}49.0 
    & \cellcolor{red!00}67.5 & \cellcolor{red!00}59.8 & \cellcolor{red!00}52.3 & \cellcolor{red!00}48.9 & \cellcolor{red!00}47.5 & \cellcolor{red!00}44.5 
    & \cellcolor{red!00}-- & \cellcolor{red!00}-- & \cellcolor{red!00}-- & \cellcolor{red!00}-- & \cellcolor{red!00}-- & \cellcolor{red!00}-- 
    \\
    TagPrime
    & \cellcolor{red!00}73.2 & \cellcolor{red!00}69.9 & \cellcolor{red!08}61.6 & \cellcolor{red!08}59.8 & \cellcolor{red!08}56.1 & \cellcolor{red!16}54.6 
    & \cellcolor{red!08}69.6 & \cellcolor{red!24}63.5 & \cellcolor{red!24}56.0 & \cellcolor{red!24}52.8 & \cellcolor{red!24}51.1 & \cellcolor{red!24}48.4 
    & \cellcolor{red!08}81.8 & \cellcolor{red!24}79.0 & \cellcolor{red!24}66.6 & \cellcolor{red!24}65.2 & \cellcolor{red!24}61.4 & \cellcolor{red!24}60.3 
    \\
    DEGREE-E2E
    & \cellcolor{red!00}70.3 & \cellcolor{red!00}66.8 & \cellcolor{red!00}57.6 & \cellcolor{red!00}55.1 & \cellcolor{red!00}51.3 & \cellcolor{red!00}49.1 
    & \cellcolor{red!00}67.7 & \cellcolor{red!00}60.5 & \cellcolor{red!00}52.2 & \cellcolor{red!00}48.7 & \cellcolor{red!00}46.6 & \cellcolor{red!00}43.7 
    & \cellcolor{red!00}74.7 & \cellcolor{red!00}70.2 & \cellcolor{red!00}38.6 & \cellcolor{red!00}33.8 & \cellcolor{red!00}25.9 & \cellcolor{red!00}23.3 
    \\
    DEGREE-PIPE
    & \cellcolor{red!00}72.0 & \cellcolor{red!00}68.4 & \cellcolor{red!00}58.6 & \cellcolor{red!08}56.3 & \cellcolor{red!00}52.9 & \cellcolor{red!00}50.7 
    & \cellcolor{red!00}68.3 & \cellcolor{red!08}61.7 & \cellcolor{red!08}52.5 & \cellcolor{red!08}48.9 & \cellcolor{red!08}47.8 & \cellcolor{red!08}44.8 
    & \cellcolor{red!00}74.0 & \cellcolor{red!00}70.4 & \cellcolor{red!00}50.9 & \cellcolor{red!08}49.6 & \cellcolor{red!08}43.6 & \cellcolor{red!08}42.7 
    \\

    \midrule
    \multirow{2}{*}{Model} & \multicolumn{6}{c|}{\textbf{Genia2011}} & \multicolumn{6}{c|}{\textbf{Genia2013}} & \multicolumn{6}{c}{\textbf{M$^2$E$^2$}} \\
    & TI & TC & AI & AC & AI+ & AC+ & TI & TC & AI & AC & AI+ & AC+ & TI & TC & AI & AC & AI+ & AC+ \\
    \midrule
    DyGIE++
    & \cellcolor{red!00}74.2 & \cellcolor{red!00}70.3 & \cellcolor{red!08}58.9 & \cellcolor{red!08}56.9 & \cellcolor{red!16}53.7 & \cellcolor{red!16}52.1 
    & \cellcolor{red!08}76.3 & \cellcolor{red!08}72.9 & \cellcolor{red!24}62.7 & \cellcolor{red!16}60.5 & \cellcolor{red!24}58.8 & \cellcolor{red!16}57.2 
    & \cellcolor{red!16}53.1 & \cellcolor{red!16}51.0 & \cellcolor{red!00}34.6 & \cellcolor{red!00}33.4 & \cellcolor{red!08}31.7 & \cellcolor{red!08}30.8 
    \\
    OneIE
    & \cellcolor{red!16}76.1 & \cellcolor{red!08}72.1 & \cellcolor{red!16}59.0 & \cellcolor{red!16}57.0 & \cellcolor{red!00}34.2 & \cellcolor{red!00}33.6 
    & \cellcolor{red!16}78.0 & \cellcolor{red!16}74.3 & \cellcolor{red!08}52.3 & \cellcolor{red!08}51.0 & \cellcolor{red!00}33.7 & \cellcolor{red!00}32.9 
    & \cellcolor{red!08}52.4 & \cellcolor{red!08}50.6 & \cellcolor{red!24}37.8 & \cellcolor{red!24}36.1 & \cellcolor{red!24}33.4 & \cellcolor{red!16}32.1 
    \\
    AMR-IE
    & \cellcolor{red!24}76.4 & \cellcolor{red!24}72.4 & \cellcolor{red!00}44.1 & \cellcolor{red!00}42.8 & \cellcolor{red!00}29.8 & \cellcolor{red!00}29.0 
    & \cellcolor{red!24}78.0 & \cellcolor{red!24}74.5 & \cellcolor{red!00}35.4 & \cellcolor{red!00}34.8 & \cellcolor{red!00}23.3 & \cellcolor{red!00}23.1 
    & \cellcolor{red!08}52.4 & \cellcolor{red!08}50.5 & \cellcolor{red!16}37.1 & \cellcolor{red!08}35.5 & \cellcolor{red!08}33.1 & \cellcolor{red!08}31.9 
    \\
    EEQA
    & \cellcolor{red!08}74.4 & \cellcolor{red!08}71.3 & \cellcolor{red!08}52.6 & \cellcolor{red!08}50.6 & \cellcolor{red!08}39.5 & \cellcolor{red!08}38.4 
    & \cellcolor{red!00}72.4 & \cellcolor{red!00}69.4 & \cellcolor{red!00}50.7 & \cellcolor{red!00}48.1 & \cellcolor{red!08}37.6 & \cellcolor{red!08}35.7 
    & \cellcolor{red!24}53.6 & \cellcolor{red!24}51.0 & \cellcolor{red!00}33.7 & \cellcolor{red!00}32.6 & \cellcolor{red!00}31.1 & \cellcolor{red!00}30.2 
    \\
    RCEE
    & \cellcolor{red!00}73.3 & \cellcolor{red!00}70.1 & \cellcolor{red!00}50.9 & \cellcolor{red!00}49.0 & \cellcolor{red!00}38.2 & \cellcolor{red!00}37.2 
    & \cellcolor{red!00}71.4 & \cellcolor{red!00}68.0 & \cellcolor{red!00}48.0 & \cellcolor{red!00}45.8 & \cellcolor{red!00}33.0 & \cellcolor{red!00}31.6 
    & \cellcolor{red!00}50.1 & \cellcolor{red!00}48.1 & \cellcolor{red!00}32.0 & \cellcolor{red!00}31.0 & \cellcolor{red!00}28.8 & \cellcolor{red!00}28.0 
    \\
    Query\&Extract
    & \cellcolor{red!00}-- & \cellcolor{red!00}-- & \cellcolor{red!00}-- & \cellcolor{red!00}-- & \cellcolor{red!00}-- & \cellcolor{red!00}-- 
    & \cellcolor{red!00}-- & \cellcolor{red!00}-- & \cellcolor{red!00}-- & \cellcolor{red!00}-- & \cellcolor{red!00}-- & \cellcolor{red!00}-- 
    & \cellcolor{red!00}51.4 & \cellcolor{red!00}49.4 & \cellcolor{red!08}35.5 & \cellcolor{red!08}33.9 & \cellcolor{red!00}30.2 & \cellcolor{red!00}28.8 
    \\
    TagPrime
    & \cellcolor{red!08}74.9 & \cellcolor{red!16}72.2 & \cellcolor{red!24}64.1 & \cellcolor{red!24}62.8 & \cellcolor{red!24}58.8 & \cellcolor{red!24}57.8 
    & \cellcolor{red!08}75.7 & \cellcolor{red!08}73.0 & \cellcolor{red!16}61.8 & \cellcolor{red!24}60.8 & \cellcolor{red!16}58.2 & \cellcolor{red!24}57.4 
    & \cellcolor{red!00}52.2 & \cellcolor{red!00}50.2 & \cellcolor{red!08}36.5 & \cellcolor{red!16}35.5 & \cellcolor{red!16}33.2 & \cellcolor{red!24}32.4 
    \\
    DEGREE-E2E
    & \cellcolor{red!00}61.6 & \cellcolor{red!00}59.2 & \cellcolor{red!00}40.0 & \cellcolor{red!00}35.6 & \cellcolor{red!00}27.7 & \cellcolor{red!00}25.4 
    & \cellcolor{red!00}66.4 & \cellcolor{red!00}62.6 & \cellcolor{red!00}37.1 & \cellcolor{red!00}33.3 & \cellcolor{red!00}27.0 & \cellcolor{red!00}24.8 
    & \cellcolor{red!00}50.9 & \cellcolor{red!00}49.5 & \cellcolor{red!00}33.7 & \cellcolor{red!00}32.5 & \cellcolor{red!00}30.9 & \cellcolor{red!00}30.0 
    \\
    DEGREE-PIPE
    & \cellcolor{red!00}63.7 & \cellcolor{red!00}60.5 & \cellcolor{red!00}51.1 & \cellcolor{red!00}49.3 & \cellcolor{red!08}40.8 & \cellcolor{red!08}39.8 
    & \cellcolor{red!00}64.9 & \cellcolor{red!00}61.0 & \cellcolor{red!08}51.0 & \cellcolor{red!08}49.4 & \cellcolor{red!08}43.0 & \cellcolor{red!08}41.9 
    & \cellcolor{red!00}50.4 & \cellcolor{red!00}48.3 & \cellcolor{red!00}34.0 & \cellcolor{red!00}33.1 & \cellcolor{red!00}30.9 & \cellcolor{red!00}30.1 
    \\

    \midrule
    \multirow{2}{*}{Model} & \multicolumn{6}{c|}{\textbf{CASIE}} & \multicolumn{6}{c|}{\textbf{PHEE}} & \multicolumn{6}{c}{--} \\
    & TI & TC & AI & AC & AI+ & AC+ & TI & TC & AI & AC & AI+ & AC+ & \multicolumn{6}{c}{--}\\
    \midrule
    DyGIE++
    & \cellcolor{red!00}44.9 & \cellcolor{red!00}44.7 & \cellcolor{red!08}37.5 & \cellcolor{red!08}36.4 & \cellcolor{red!08}30.4 & \cellcolor{red!08}29.5 
    & \cellcolor{red!08}71.4 & \cellcolor{red!08}70.4 & \cellcolor{red!24}69.9 & \cellcolor{red!24}60.8 & \cellcolor{red!24}52.4 & \cellcolor{red!24}45.7 
    \\
    OneIE
    & \cellcolor{red!16}70.8 & \cellcolor{red!16}70.6 & \cellcolor{red!16}57.2 & \cellcolor{red!16}54.2 & \cellcolor{red!00}23.1 & \cellcolor{red!00}22.1 
    & \cellcolor{red!00}70.9 & \cellcolor{red!00}70.0 & \cellcolor{red!00}51.5 & \cellcolor{red!00}37.5 & \cellcolor{red!00}40.1 & \cellcolor{red!00}29.8 
    \\
    AMR-IE
    & \cellcolor{red!24}71.1 & \cellcolor{red!24}70.8 & \cellcolor{red!00}34.5 & \cellcolor{red!00}10.7 & \cellcolor{red!00}10.0 & \cellcolor{red!00}3.1 
    & \cellcolor{red!00}70.2 & \cellcolor{red!00}69.4 & \cellcolor{red!00}57.1 & \cellcolor{red!00}45.7 & \cellcolor{red!00}42.2 & \cellcolor{red!00}34.1 
    \\
    EEQA
    & \cellcolor{red!00}43.2 & \cellcolor{red!00}42.8 & \cellcolor{red!00}36.2 & \cellcolor{red!00}35.1 & \cellcolor{red!08}27.0 & \cellcolor{red!08}26.2 
    & \cellcolor{red!08}70.9 & \cellcolor{red!08}70.3 & \cellcolor{red!00}48.5 & \cellcolor{red!00}40.4 & \cellcolor{red!00}38.1 & \cellcolor{red!00}32.0 
    \\
    RCEE
    & \cellcolor{red!00}42.3 & \cellcolor{red!00}42.1 & \cellcolor{red!00}34.1 & \cellcolor{red!00}32.8 & \cellcolor{red!00}24.6 & \cellcolor{red!00}23.7 
    & \cellcolor{red!16}71.6 & \cellcolor{red!16}70.9 & \cellcolor{red!00}49.1 & \cellcolor{red!00}41.6 & \cellcolor{red!00}38.7 & \cellcolor{red!00}33.1 
    \\
    Query\&Extract
    & \cellcolor{red!00}-- & \cellcolor{red!00}-- & \cellcolor{red!00}-- & \cellcolor{red!00}-- & \cellcolor{red!00}-- & \cellcolor{red!00}-- 
    & \cellcolor{red!00}66.2 & \cellcolor{red!00}55.5 & \cellcolor{red!00}48.1 & \cellcolor{red!00}41.4 & \cellcolor{red!00}36.7 & \cellcolor{red!00}31.8 
    \\
    TagPrime
    & \cellcolor{red!08}69.5 & \cellcolor{red!08}69.3 & \cellcolor{red!24}63.3 & \cellcolor{red!24}61.0 & \cellcolor{red!24}50.9 & \cellcolor{red!24}49.1 
    & \cellcolor{red!24}71.7 & \cellcolor{red!24}71.1 & \cellcolor{red!16}60.9 & \cellcolor{red!16}51.7 & \cellcolor{red!16}47.4 & \cellcolor{red!16}40.6 
    \\
    DEGREE-E2E
    & \cellcolor{red!08}60.9 & \cellcolor{red!08}60.7 & \cellcolor{red!00}36.0 & \cellcolor{red!00}27.0 & \cellcolor{red!00}18.5 & \cellcolor{red!00}14.6 
    & \cellcolor{red!00}70.0 & \cellcolor{red!00}69.1 & \cellcolor{red!08}57.5 & \cellcolor{red!08}49.3 & \cellcolor{red!08}42.4 & \cellcolor{red!08}36.5 
    \\
    DEGREE-PIPE
    & \cellcolor{red!00}57.4 & \cellcolor{red!00}57.1 & \cellcolor{red!08}49.7 & \cellcolor{red!08}48.0 & \cellcolor{red!16}34.8 & \cellcolor{red!16}33.7 
    & \cellcolor{red!00}69.8 & \cellcolor{red!00}69.1 & \cellcolor{red!08}59.0 & \cellcolor{red!08}50.2 & \cellcolor{red!08}42.8 & \cellcolor{red!08}36.7 
    \\
    
    \bottomrule

\end{tabular}}
\caption{Reevaluation results for end-to-end event extraction (E2E). All the numbers are the average score of 5 data splits. Darker cells imply higher scores. We use ``--'' to denote the cases that models are not runnable.}
\label{tab:all_e2e}
\end{table*}

\begin{table*}[t!]
\centering
\setlength{\tabcolsep}{7pt}
\resizebox{.77\textwidth}{!}{
\begin{tabular}{l|cc|cc|cc|cc|cc|cc}
    \toprule
    \multirow{2}{*}{Model} & \multicolumn{2}{c|}{\textbf{ACE05}} & \multicolumn{2}{c|}{\textbf{RichERE}} & \multicolumn{2}{c|}{\textbf{MLEE}} & \multicolumn{2}{c|}{\textbf{Genia2011}} & \multicolumn{2}{c|}{\textbf{Genia2013}} & \multicolumn{2}{c}{\textbf{M$^2$E$^2$}} \\
    & TI & TC & TI & TC & TI & TC & TI & TC & TI & TC & TI & TC \\
    \midrule
    DyGIE++
    & \cellcolor{red!16}74.7 & \cellcolor{red!24}71.3 
    & \cellcolor{red!16}69.7 & \cellcolor{red!00}59.8 
    & \cellcolor{red!16}82.6 & \cellcolor{red!08}78.2 
    & \cellcolor{red!08}74.2 & \cellcolor{red!08}70.3 
    & \cellcolor{red!16}76.3 & \cellcolor{red!08}72.9 
    & \cellcolor{red!16}53.1 & \cellcolor{red!16}51.0 
    \\
    OneIE
    & \cellcolor{red!24}75.0 & \cellcolor{red!16}71.1 
    & \cellcolor{red!24}71.0 & \cellcolor{red!16}62.5 
    & \cellcolor{red!24}82.7 & \cellcolor{red!16}78.5 
    & \cellcolor{red!16}76.1 & \cellcolor{red!16}72.1 
    & \cellcolor{red!16}78.0 & \cellcolor{red!16}74.3 
    & \cellcolor{red!08}52.4 & \cellcolor{red!16}50.6 
    \\
    AMR-IE
    & \cellcolor{red!16}74.6 & \cellcolor{red!16}71.1 
    & \cellcolor{red!16}70.5 & \cellcolor{red!16}62.3 
    & \cellcolor{red!16}82.4 & \cellcolor{red!16}78.2 
    & \cellcolor{red!24}76.4 & \cellcolor{red!24}72.4 
    & \cellcolor{red!24}78.0 & \cellcolor{red!24}74.5 
    & \cellcolor{red!16}52.4 & \cellcolor{red!08}50.5 
    \\
    EEQA
    & \cellcolor{red!00}73.8 & \cellcolor{red!08}70.0 
    & \cellcolor{red!00}69.3 & \cellcolor{red!00}60.2 
    & \cellcolor{red!08}82.0 & \cellcolor{red!08}77.4 
    & \cellcolor{red!00}73.3 & \cellcolor{red!00}69.6 
    & \cellcolor{red!08}74.7 & \cellcolor{red!08}71.1 
    & \cellcolor{red!24}53.6 & \cellcolor{red!24}51.0 
    \\
    RCEE
    & \cellcolor{red!08}74.0 & \cellcolor{red!08}70.5 
    & \cellcolor{red!00}68.6 & \cellcolor{red!00}60.0 
    & \cellcolor{red!08}82.0 & \cellcolor{red!00}77.3 
    & \cellcolor{red!00}73.1 & \cellcolor{red!00}69.3 
    & \cellcolor{red!00}74.6 & \cellcolor{red!00}70.8 
    & \cellcolor{red!00}50.1 & \cellcolor{red!00}48.1 
    \\
    Query\&Extract
    & \cellcolor{red!00}68.6 & \cellcolor{red!00}65.1 
    & \cellcolor{red!00}67.5 & \cellcolor{red!00}59.8 
    & \cellcolor{red!00}78.0 & \cellcolor{red!00}74.9 
    & \cellcolor{red!00}71.6 & \cellcolor{red!00}68.9 
    & \cellcolor{red!00}73.0 & \cellcolor{red!00}70.1 
    & \cellcolor{red!00}51.4 & \cellcolor{red!00}49.4 
    \\
    TagPrime-C
    & \cellcolor{red!00}73.2 & \cellcolor{red!00}69.9 
    & \cellcolor{red!08}69.6 & \cellcolor{red!24}63.5 
    & \cellcolor{red!00}81.8 & \cellcolor{red!24}79.0 
    & \cellcolor{red!16}74.9 & \cellcolor{red!16}72.2 
    & \cellcolor{red!08}75.7 & \cellcolor{red!16}73.0 
    & \cellcolor{red!08}52.2 & \cellcolor{red!08}50.2 
    \\
    UniST
    & \cellcolor{red!08}73.9 & \cellcolor{red!00}69.8 
    & \cellcolor{red!08}69.6 & \cellcolor{red!08}60.7 
    & \cellcolor{red!00}80.2 & \cellcolor{red!00}74.9 
    & \cellcolor{red!08}73.8 & \cellcolor{red!08}70.3 
    & \cellcolor{red!00}73.7 & \cellcolor{red!00}69.9 
    & \cellcolor{red!00}51.1 & \cellcolor{red!00}49.0 
    \\
    CEDAR
    & \cellcolor{red!00}71.9 & \cellcolor{red!00}62.6 
    & \cellcolor{red!00}67.3 & \cellcolor{red!00}52.3 
    & \cellcolor{red!00}71.0 & \cellcolor{red!00}65.5 
    & \cellcolor{red!00}70.2 & \cellcolor{red!00}66.8 
    & \cellcolor{red!00}73.6 & \cellcolor{red!00}67.1 
    & \cellcolor{red!00}50.9 & \cellcolor{red!00}48.0 
    \\
    DEGREE
    & \cellcolor{red!00}72.0 & \cellcolor{red!00}68.4 
    & \cellcolor{red!00}68.3 & \cellcolor{red!08}61.7 
    & \cellcolor{red!00}74.0 & \cellcolor{red!00}70.4 
    & \cellcolor{red!00}63.7 & \cellcolor{red!00}60.5 
    & \cellcolor{red!00}64.9 & \cellcolor{red!00}61.0 
    & \cellcolor{red!00}50.4 & \cellcolor{red!00}48.3 
    \\

    \midrule
    \multirow{2}{*}{Model} & \multicolumn{2}{c|}{\textbf{CASIE}} & \multicolumn{2}{c|}{\textbf{PHEE}} & \multicolumn{2}{c|}{\textbf{MAVEN}} & \multicolumn{2}{c|}{\textbf{FewEvent}} & \multicolumn{2}{c|}{\textbf{MEE-en}} & \multicolumn{2}{c}{\textbf{SPEED}} \\
    & TI & TC & TI & TC & TI & TC & TI & TC & TI & TC & TI & TC \\
    \midrule
        DyGIE++
    & \cellcolor{red!00}44.9 & \cellcolor{red!00}44.7 
    & \cellcolor{red!16}71.4 & \cellcolor{red!16}70.4 
    & \cellcolor{red!08}75.9 & \cellcolor{red!08}65.3 
    & \cellcolor{red!16}67.7 & \cellcolor{red!08}65.2 
    & \cellcolor{red!24}81.7 & \cellcolor{red!16}79.8 
    & \cellcolor{red!08}69.6 & \cellcolor{red!00}64.9 
    \\
    OneIE
    & \cellcolor{red!16}70.8 & \cellcolor{red!16}70.6 
    & \cellcolor{red!00}70.9 & \cellcolor{red!00}70.0 
    & \cellcolor{red!16}76.4 & \cellcolor{red!16}65.5 
    & \cellcolor{red!08}67.5 & \cellcolor{red!16}65.4 
    & \cellcolor{red!00}80.7 & \cellcolor{red!08}78.8 
    & \cellcolor{red!00}69.5 & \cellcolor{red!08}65.1 
    \\
    AMR-IE
    & \cellcolor{red!24}71.1 & \cellcolor{red!24}70.8 
    & \cellcolor{red!00}70.2 & \cellcolor{red!00}69.4 
    & \cellcolor{red!00}-- & \cellcolor{red!00}-- 
    & \cellcolor{red!08}67.4 & \cellcolor{red!08}65.2 
    & \cellcolor{red!00}-- & \cellcolor{red!00}-- 
    & \cellcolor{red!00}-- & \cellcolor{red!00}-- 
    \\
    EEQA
    & \cellcolor{red!00}43.4 & \cellcolor{red!00}43.2 
    & \cellcolor{red!08}70.9 & \cellcolor{red!08}70.3 
    & \cellcolor{red!00}75.2 & \cellcolor{red!00}64.4 
    & \cellcolor{red!00}67.0 & \cellcolor{red!00}65.1 
    & \cellcolor{red!08}81.4 & \cellcolor{red!16}79.5 
    & \cellcolor{red!08}69.9 & \cellcolor{red!16}65.3 
    \\
    RCEE
    & \cellcolor{red!00}43.5 & \cellcolor{red!00}43.3 
    & \cellcolor{red!16}71.6 & \cellcolor{red!16}70.9 
    & \cellcolor{red!00}75.2 & \cellcolor{red!08}64.6 
    & \cellcolor{red!00}67.0 & \cellcolor{red!00}65.0 
    & \cellcolor{red!08}81.1 & \cellcolor{red!08}79.1 
    & \cellcolor{red!16}70.1 & \cellcolor{red!08}65.1 
    \\
    Query\&Extract
    & \cellcolor{red!00}51.6 & \cellcolor{red!00}51.5 
    & \cellcolor{red!00}66.2 & \cellcolor{red!00}55.5 
    & \cellcolor{red!00}-- & \cellcolor{red!00}-- 
    & \cellcolor{red!00}66.3 & \cellcolor{red!00}63.8 
    & \cellcolor{red!00}80.2 & \cellcolor{red!00}78.1 
    & \cellcolor{red!16}70.2 & \cellcolor{red!16}66.2 
    \\
    TagPrime-C
    & \cellcolor{red!16}69.5 & \cellcolor{red!16}69.3 
    & \cellcolor{red!24}71.7 & \cellcolor{red!24}71.1 
    & \cellcolor{red!00}74.7 & \cellcolor{red!24}66.1 
    & \cellcolor{red!00}67.2 & \cellcolor{red!24}65.6 
    & \cellcolor{red!16}81.5 & \cellcolor{red!24}79.8 
    & \cellcolor{red!24}70.3 & \cellcolor{red!24}66.4 
    \\
    UniST
    & \cellcolor{red!08}68.4 & \cellcolor{red!08}68.1 
    & \cellcolor{red!00}70.7 & \cellcolor{red!00}69.6 
    & \cellcolor{red!24}76.7 & \cellcolor{red!00}63.4 
    & \cellcolor{red!16}67.5 & \cellcolor{red!00}63.1 
    & \cellcolor{red!00}80.5 & \cellcolor{red!00}78.3 
    & \cellcolor{red!00}-- & \cellcolor{red!00}-- 
    \\
    CEDAR
    & \cellcolor{red!08}68.7 & \cellcolor{red!08}67.6 
    & \cellcolor{red!08}71.2 & \cellcolor{red!08}70.3 
    & \cellcolor{red!16}76.5 & \cellcolor{red!00}54.5 
    & \cellcolor{red!00}66.9 & \cellcolor{red!00}52.1 
    & \cellcolor{red!16}81.5 & \cellcolor{red!00}78.6 
    & \cellcolor{red!00}67.6 & \cellcolor{red!00}61.7 
    \\
    DEGREE
    & \cellcolor{red!00}61.5 & \cellcolor{red!00}61.3 
    & \cellcolor{red!00}69.8 & \cellcolor{red!00}69.1 
    & \cellcolor{red!08}76.2 & \cellcolor{red!16}65.5 
    & \cellcolor{red!24}67.9 & \cellcolor{red!16}65.5 
    & \cellcolor{red!00}80.2 & \cellcolor{red!00}78.2 
    & \cellcolor{red!00}66.5 & \cellcolor{red!00}62.2 
    \\
    \bottomrule

\end{tabular}}
\caption{Reevaluation results for event detection (ED). All the numbers are the average score of 5 data splits. Darker cells imply higher scores. We use ``--'' to denote the cases that models are not runnable.}
\label{tab:all_ed}
\end{table*}

\begin{table*}[t!]
\centering
\setlength{\tabcolsep}{5.5pt}
\resizebox{.99\textwidth}{!}{
\begin{tabular}{l|cccc|cccc|cccc|cccc}
    \toprule
    \multirow{2}{*}{Model} & \multicolumn{4}{c|}{\textbf{ACE05}} & \multicolumn{4}{c|}{\textbf{RichERE}} & \multicolumn{4}{c|}{\textbf{MLEE}} & \multicolumn{4}{c}{\textbf{Genia2011}} \\
    & AI & AC & AI+ & AC+ & AI & AC & AI+ & AC+ & AI & AC & AI+ & AC+ & AI & AC & AI+ & AC+ \\
    \midrule
    DyGIE++
    & \cellcolor{red!00}66.9 & \cellcolor{red!00}61.5 & \cellcolor{red!00}65.2 & \cellcolor{red!00}60.0 
    & \cellcolor{red!00}58.5 & \cellcolor{red!00}49.4 & \cellcolor{red!00}56.2 & \cellcolor{red!00}47.3 
    & \cellcolor{red!08}67.9 & \cellcolor{red!00}64.8 & \cellcolor{red!08}65.2 & \cellcolor{red!08}62.4 
    & \cellcolor{red!00}66.1 & \cellcolor{red!00}63.7 & \cellcolor{red!00}63.0 & \cellcolor{red!00}61.0 
    \\
    OneIE
    & \cellcolor{red!00}75.4 & \cellcolor{red!00}71.5 & \cellcolor{red!00}74.0 & \cellcolor{red!00}70.2 
    & \cellcolor{red!00}71.6 & \cellcolor{red!00}65.8 & \cellcolor{red!00}69.3 & \cellcolor{red!00}63.7 
    & \cellcolor{red!00}31.0 & \cellcolor{red!00}28.9 & \cellcolor{red!00}16.4 & \cellcolor{red!00}15.7 
    & \cellcolor{red!00}62.9 & \cellcolor{red!00}60.3 & \cellcolor{red!00}40.1 & \cellcolor{red!00}38.9 
    \\
    AMR-IE
    & \cellcolor{red!08}76.2 & \cellcolor{red!00}72.6 & \cellcolor{red!00}74.5 & \cellcolor{red!00}70.9 
    & \cellcolor{red!00}72.8 & \cellcolor{red!00}65.8 & \cellcolor{red!00}69.6 & \cellcolor{red!00}63.0 
    & \cellcolor{red!00}23.2 & \cellcolor{red!00}16.6 & \cellcolor{red!00}8.0 & \cellcolor{red!00}6.1 
    & \cellcolor{red!00}49.1 & \cellcolor{red!00}47.6 & \cellcolor{red!00}36.1 & \cellcolor{red!00}35.3 
    \\
    EEQA
    & \cellcolor{red!00}73.8 & \cellcolor{red!00}71.4 & \cellcolor{red!00}71.9 & \cellcolor{red!00}69.6 
    & \cellcolor{red!00}73.3 & \cellcolor{red!00}67.3 & \cellcolor{red!00}70.8 & \cellcolor{red!00}64.9 
    & \cellcolor{red!00}64.8 & \cellcolor{red!00}62.1 & \cellcolor{red!00}51.4 & \cellcolor{red!00}49.5 
    & \cellcolor{red!00}63.2 & \cellcolor{red!00}60.8 & \cellcolor{red!00}51.2 & \cellcolor{red!00}49.4 
    \\
    RCEE
    & \cellcolor{red!00}73.7 & \cellcolor{red!00}71.2 & \cellcolor{red!00}71.8 & \cellcolor{red!00}69.4 
    & \cellcolor{red!00}72.8 & \cellcolor{red!00}67.0 & \cellcolor{red!00}70.2 & \cellcolor{red!00}64.5 
    & \cellcolor{red!00}61.1 & \cellcolor{red!00}58.2 & \cellcolor{red!00}47.3 & \cellcolor{red!00}45.1 
    & \cellcolor{red!00}62.3 & \cellcolor{red!00}59.9 & \cellcolor{red!00}51.4 & \cellcolor{red!00}49.6 
    \\
    Query\&Extract
    & \cellcolor{red!16}77.3 & \cellcolor{red!16}73.6 & \cellcolor{red!16}75.7 & \cellcolor{red!16}72.0 
    & \cellcolor{red!16}76.4 & \cellcolor{red!16}70.9 & \cellcolor{red!16}74.7 & \cellcolor{red!16}69.2 
    & \cellcolor{red!00}-- & \cellcolor{red!00}-- & \cellcolor{red!00}-- & \cellcolor{red!00}-- 
    & \cellcolor{red!00}-- & \cellcolor{red!00}-- & \cellcolor{red!00}-- & \cellcolor{red!00}-- 
    \\
    TagPrime-C
    & \cellcolor{red!24}80.0 & \cellcolor{red!24}76.0 & \cellcolor{red!24}78.5 & \cellcolor{red!24}74.5 
    & \cellcolor{red!24}78.8 & \cellcolor{red!24}73.3 & \cellcolor{red!24}76.7 & \cellcolor{red!24}71.4 
    & \cellcolor{red!24}78.9 & \cellcolor{red!24}76.6 & \cellcolor{red!24}76.5 & \cellcolor{red!24}74.5 
    & \cellcolor{red!24}79.6 & \cellcolor{red!24}77.4 & \cellcolor{red!24}77.7 & \cellcolor{red!24}75.8 
    \\
    TagPrime-CR
    & \cellcolor{red!24}80.1 & \cellcolor{red!24}77.8 & \cellcolor{red!24}78.5 & \cellcolor{red!24}76.2 
    & \cellcolor{red!24}78.7 & \cellcolor{red!24}74.3 & \cellcolor{red!24}76.6 & \cellcolor{red!24}72.5 
    & \cellcolor{red!24}79.2 & \cellcolor{red!24}77.3 & \cellcolor{red!24}76.4 & \cellcolor{red!24}74.6 
    & \cellcolor{red!24}78.0 & \cellcolor{red!24}76.2 & \cellcolor{red!24}76.2 & \cellcolor{red!24}74.5 
    \\
    DEGREE
    & \cellcolor{red!08}76.4 & \cellcolor{red!08}73.3 & \cellcolor{red!08}74.9 & \cellcolor{red!08}71.8 
    & \cellcolor{red!08}75.1 & \cellcolor{red!08}70.2 & \cellcolor{red!08}73.6 & \cellcolor{red!08}68.8 
    & \cellcolor{red!00}67.6 & \cellcolor{red!08}65.3 & \cellcolor{red!00}63.4 & \cellcolor{red!00}61.5 
    & \cellcolor{red!00}68.2 & \cellcolor{red!00}65.7 & \cellcolor{red!00}64.5 & \cellcolor{red!00}62.4 
    \\
    BART-Gen
    & \cellcolor{red!00}76.0 & \cellcolor{red!08}72.6 & \cellcolor{red!08}74.8 & \cellcolor{red!08}71.2 
    & \cellcolor{red!00}74.4 & \cellcolor{red!00}68.8 & \cellcolor{red!00}73.1 & \cellcolor{red!00}67.7 
    & \cellcolor{red!16}73.1 & \cellcolor{red!16}69.8 & \cellcolor{red!16}71.8 & \cellcolor{red!16}68.7 
    & \cellcolor{red!16}73.4 & \cellcolor{red!16}70.9 & \cellcolor{red!16}71.8 & \cellcolor{red!16}69.5 
    \\
    X-Gear
    & \cellcolor{red!00}76.1 & \cellcolor{red!00}72.4 & \cellcolor{red!00}74.4 & \cellcolor{red!00}70.8 
    & \cellcolor{red!08}75.0 & \cellcolor{red!00}68.7 & \cellcolor{red!08}73.4 & \cellcolor{red!00}67.2 
    & \cellcolor{red!00}64.8 & \cellcolor{red!00}63.3 & \cellcolor{red!00}60.7 & \cellcolor{red!00}59.4 
    & \cellcolor{red!08}68.4 & \cellcolor{red!08}66.2 & \cellcolor{red!08}65.0 & \cellcolor{red!08}63.1 
    \\
    PAIE
    & \cellcolor{red!16}77.2 & \cellcolor{red!16}74.0 & \cellcolor{red!16}76.0 & \cellcolor{red!16}72.9 
    & \cellcolor{red!16}76.6 & \cellcolor{red!16}71.1 & \cellcolor{red!16}75.3 & \cellcolor{red!16}70.0 
    & \cellcolor{red!16}76.0 & \cellcolor{red!16}73.5 & \cellcolor{red!16}74.7 & \cellcolor{red!16}72.4 
    & \cellcolor{red!16}76.8 & \cellcolor{red!16}74.6 & \cellcolor{red!16}75.5 & \cellcolor{red!16}73.4 
    \\
    Ampere
    & \cellcolor{red!00}75.5 & \cellcolor{red!00}72.0 & \cellcolor{red!00}73.9 & \cellcolor{red!00}70.6 
    & \cellcolor{red!00}73.8 & \cellcolor{red!08}69.2 & \cellcolor{red!00}72.2 & \cellcolor{red!08}67.7 
    & \cellcolor{red!08}69.2 & \cellcolor{red!08}67.1 & \cellcolor{red!08}64.4 & \cellcolor{red!08}62.6 
    & \cellcolor{red!08}69.5 & \cellcolor{red!08}67.1 & \cellcolor{red!08}66.0 & \cellcolor{red!08}63.8 
    \\
    
    \midrule
    \multirow{2}{*}{Model} & \multicolumn{4}{c|}{\textbf{Genia2013}} & \multicolumn{4}{c|}{\textbf{M$^2$E$^2$}} & \multicolumn{4}{c|}{\textbf{CASIE}} & \multicolumn{4}{c}{\textbf{PHEE}} \\
    & AI & AC & AI+ & AC+ & AI & AC & AI+ & AC+ & AI & AC & AI+ & AC+ & AI & AC & AI+ & AC+ \\
    \midrule
    DyGIE++
    & \cellcolor{red!08}71.7 & \cellcolor{red!08}69.3 & \cellcolor{red!08}68.7 & \cellcolor{red!08}66.9 
    & \cellcolor{red!00}41.7 & \cellcolor{red!00}38.9 & \cellcolor{red!00}41.0 & \cellcolor{red!00}38.5 
    & \cellcolor{red!00}58.0 & \cellcolor{red!00}56.0 & \cellcolor{red!00}53.4 & \cellcolor{red!00}51.5 
    & \cellcolor{red!08}63.4 & \cellcolor{red!08}54.6 & \cellcolor{red!08}63.0 & \cellcolor{red!08}54.2 
    \\
    OneIE
    & \cellcolor{red!00}57.2 & \cellcolor{red!00}55.7 & \cellcolor{red!00}39.4 & \cellcolor{red!00}38.7 
    & \cellcolor{red!00}59.0 & \cellcolor{red!00}55.2 & \cellcolor{red!00}57.2 & \cellcolor{red!00}53.3 
    & \cellcolor{red!00}58.3 & \cellcolor{red!00}55.3 & \cellcolor{red!00}29.0 & \cellcolor{red!00}27.7 
    & \cellcolor{red!00}55.9 & \cellcolor{red!00}40.6 & \cellcolor{red!00}55.5 & \cellcolor{red!00}40.4 
    \\
    AMR-IE
    & \cellcolor{red!00}38.9 & \cellcolor{red!00}38.1 & \cellcolor{red!00}26.7 & \cellcolor{red!00}26.4 
    & \cellcolor{red!00}56.0 & \cellcolor{red!00}51.3 & \cellcolor{red!00}55.3 & \cellcolor{red!00}50.4 
    & \cellcolor{red!00}35.5 & \cellcolor{red!00}11.0 & \cellcolor{red!00}12.8 & \cellcolor{red!00}4.0 
    & \cellcolor{red!00}60.4 & \cellcolor{red!00}45.3 & \cellcolor{red!00}59.9 & \cellcolor{red!00}44.9 
    \\
    EEQA
    & \cellcolor{red!00}64.7 & \cellcolor{red!00}61.1 & \cellcolor{red!00}50.3 & \cellcolor{red!00}47.5 
    & \cellcolor{red!00}57.6 & \cellcolor{red!00}55.9 & \cellcolor{red!00}57.0 & \cellcolor{red!00}55.3 
    & \cellcolor{red!00}56.1 & \cellcolor{red!00}54.0 & \cellcolor{red!00}50.9 & \cellcolor{red!00}49.0 
    & \cellcolor{red!00}53.7 & \cellcolor{red!00}45.6 & \cellcolor{red!00}53.4 & \cellcolor{red!00}45.4 
    \\
    RCEE
    & \cellcolor{red!00}60.7 & \cellcolor{red!00}57.4 & \cellcolor{red!00}45.1 & \cellcolor{red!00}42.7 
    & \cellcolor{red!00}57.9 & \cellcolor{red!00}56.4 & \cellcolor{red!00}57.3 & \cellcolor{red!00}55.8 
    & \cellcolor{red!00}47.6 & \cellcolor{red!00}45.3 & \cellcolor{red!00}41.5 & \cellcolor{red!00}39.5 
    & \cellcolor{red!00}54.1 & \cellcolor{red!00}45.8 & \cellcolor{red!00}53.8 & \cellcolor{red!00}45.6 
    \\
    Query\&Extract
    & \cellcolor{red!00}-- & \cellcolor{red!00}-- & \cellcolor{red!00}-- & \cellcolor{red!00}-- 
    & \cellcolor{red!00}59.9 & \cellcolor{red!00}56.2 & \cellcolor{red!00}58.0 & \cellcolor{red!00}54.2 
    & \cellcolor{red!00}-- & \cellcolor{red!00}-- & \cellcolor{red!00}-- & \cellcolor{red!00}-- 
    & \cellcolor{red!08}64.6 & \cellcolor{red!08}54.8 & \cellcolor{red!08}64.2 & \cellcolor{red!08}54.4 
    \\
    TagPrime-C
    & \cellcolor{red!24}79.8 & \cellcolor{red!24}77.4 & \cellcolor{red!24}77.1 & \cellcolor{red!24}74.9 
    & \cellcolor{red!24}63.4 & \cellcolor{red!16}60.1 & \cellcolor{red!16}62.3 & \cellcolor{red!08}59.0 
    & \cellcolor{red!24}71.9 & \cellcolor{red!24}69.1 & \cellcolor{red!24}68.8 & \cellcolor{red!24}66.1 
    & \cellcolor{red!16}66.0 & \cellcolor{red!16}55.6 & \cellcolor{red!16}65.6 & \cellcolor{red!16}55.3 
    \\
    TagPrime-CR
    & \cellcolor{red!16}76.6 & \cellcolor{red!16}74.5 & \cellcolor{red!16}74.3 & \cellcolor{red!16}72.3 
    & \cellcolor{red!24}63.2 & \cellcolor{red!24}60.8 & \cellcolor{red!24}62.3 & \cellcolor{red!24}59.9 
    & \cellcolor{red!24}71.1 & \cellcolor{red!24}69.2 & \cellcolor{red!24}67.9 & \cellcolor{red!24}66.1 
    & \cellcolor{red!16}65.8 & \cellcolor{red!16}56.0 & \cellcolor{red!16}65.5 & \cellcolor{red!16}55.7 
    \\
    DEGREE
    & \cellcolor{red!00}68.4 & \cellcolor{red!00}66.0 & \cellcolor{red!00}64.6 & \cellcolor{red!00}62.5 
    & \cellcolor{red!08}62.3 & \cellcolor{red!08}59.8 & \cellcolor{red!08}61.7 & \cellcolor{red!16}59.2 
    & \cellcolor{red!00}61.0 & \cellcolor{red!08}59.0 & \cellcolor{red!08}56.5 & \cellcolor{red!08}54.7 
    & \cellcolor{red!00}61.7 & \cellcolor{red!00}52.5 & \cellcolor{red!00}61.4 & \cellcolor{red!00}52.3 
    \\
    BART-Gen
    & \cellcolor{red!16}76.4 & \cellcolor{red!16}73.6 & \cellcolor{red!16}74.8 & \cellcolor{red!16}72.2 
    & \cellcolor{red!08}62.5 & \cellcolor{red!16}60.0 & \cellcolor{red!16}62.1 & \cellcolor{red!16}59.6 
    & \cellcolor{red!08}63.7 & \cellcolor{red!08}60.0 & \cellcolor{red!16}61.8 & \cellcolor{red!08}58.3 
    & \cellcolor{red!00}57.1 & \cellcolor{red!00}47.7 & \cellcolor{red!00}56.9 & \cellcolor{red!00}47.5 
    \\
    X-Gear
    & \cellcolor{red!00}64.1 & \cellcolor{red!00}61.9 & \cellcolor{red!00}60.5 & \cellcolor{red!00}58.6 
    & \cellcolor{red!16}62.7 & \cellcolor{red!08}59.8 & \cellcolor{red!08}61.9 & \cellcolor{red!08}59.0 
    & \cellcolor{red!16}65.7 & \cellcolor{red!16}63.4 & \cellcolor{red!08}61.4 & \cellcolor{red!16}59.3 
    & \cellcolor{red!24}67.6 & \cellcolor{red!24}58.3 & \cellcolor{red!24}67.4 & \cellcolor{red!24}58.2 
    \\
    PAIE
    & \cellcolor{red!24}77.8 & \cellcolor{red!24}75.2 & \cellcolor{red!24}76.6 & \cellcolor{red!24}74.2 
    & \cellcolor{red!16}62.9 & \cellcolor{red!24}60.6 & \cellcolor{red!24}62.7 & \cellcolor{red!24}60.4 
    & \cellcolor{red!16}68.1 & \cellcolor{red!16}65.7 & \cellcolor{red!16}66.4 & \cellcolor{red!16}64.0 
    & \cellcolor{red!24}74.9 & \cellcolor{red!24}73.3 & \cellcolor{red!24}74.7 & \cellcolor{red!24}73.1 
    \\
    Ampere
    & \cellcolor{red!08}73.2 & \cellcolor{red!08}71.0 & \cellcolor{red!08}69.6 & \cellcolor{red!08}67.7 
    & \cellcolor{red!00}62.1 & \cellcolor{red!00}59.1 & \cellcolor{red!00}61.4 & \cellcolor{red!00}58.4 
    & \cellcolor{red!08}61.1 & \cellcolor{red!00}58.4 & \cellcolor{red!00}56.4 & \cellcolor{red!00}53.9 
    & \cellcolor{red!00}61.4 & \cellcolor{red!00}51.7 & \cellcolor{red!00}61.1 & \cellcolor{red!00}51.6 
    \\

    \midrule
    \multirow{2}{*}{Model} & \multicolumn{4}{c|}{\textbf{WikiEvnts}} & \multicolumn{4}{c|}{\textbf{RAMS}} & \multicolumn{4}{c|}{\textbf{GENEVA}} & \multicolumn{4}{c}{\textbf{MUC-4}} \\
    & AI & AC & AI+ & AC+ & AI & AC & AI+ & AC+ & AI & AC & AI+ & AC+ & AI & AC & AI+ & AC+ \\
    \midrule
    DyGIE++
    & \cellcolor{red!00}39.8 & \cellcolor{red!00}35.3 & \cellcolor{red!00}39.0 & \cellcolor{red!00}34.7 
    & \cellcolor{red!00}44.3 & \cellcolor{red!00}35.3 & \cellcolor{red!00}44.3 & \cellcolor{red!00}35.3 
    & \cellcolor{red!00}66.0 & \cellcolor{red!00}62.5 & \cellcolor{red!00}65.8 & \cellcolor{red!00}62.3 
    & \cellcolor{red!24}56.5 & \cellcolor{red!24}55.6 & \cellcolor{red!24}56.5 & \cellcolor{red!24}55.6 
    \\
    OneIE
    & \cellcolor{red!00}17.5 & \cellcolor{red!00}15.0 & \cellcolor{red!00}9.2 & \cellcolor{red!00}7.9 
    & \cellcolor{red!00}48.0 & \cellcolor{red!00}40.7 & \cellcolor{red!00}48.0 & \cellcolor{red!00}40.7 
    & \cellcolor{red!00}38.9 & \cellcolor{red!00}37.1 & \cellcolor{red!00}38.6 & \cellcolor{red!00}36.9 
    & \cellcolor{red!16}55.1 & \cellcolor{red!16}53.9 & \cellcolor{red!16}55.1 & \cellcolor{red!16}53.9 
    \\
    AMR-IE
    & \cellcolor{red!00}17.8 & \cellcolor{red!00}16.0 & \cellcolor{red!00}11.7 & \cellcolor{red!00}10.4 
    & \cellcolor{red!00}49.6 & \cellcolor{red!00}42.3 & \cellcolor{red!00}49.6 & \cellcolor{red!00}42.3 
    & \cellcolor{red!00}23.7 & \cellcolor{red!00}16.6 & \cellcolor{red!00}23.4 & \cellcolor{red!00}16.4 
    & \cellcolor{red!00}-- & \cellcolor{red!00}-- & \cellcolor{red!00}-- & \cellcolor{red!00}-- 
    \\
    EEQA
    & \cellcolor{red!00}54.3 & \cellcolor{red!00}51.7 & \cellcolor{red!00}48.4 & \cellcolor{red!00}46.1 
    & \cellcolor{red!00}48.9 & \cellcolor{red!00}44.7 & \cellcolor{red!00}48.9 & \cellcolor{red!00}44.7 
    & \cellcolor{red!08}69.7 & \cellcolor{red!08}67.3 & \cellcolor{red!08}69.4 & \cellcolor{red!08}67.0 
    & \cellcolor{red!00}32.7 & \cellcolor{red!00}27.4 & \cellcolor{red!00}32.7 & \cellcolor{red!00}27.4 
    \\
    RCEE
    & \cellcolor{red!00}53.7 & \cellcolor{red!00}50.9 & \cellcolor{red!00}46.4 & \cellcolor{red!00}44.0 
    & \cellcolor{red!00}45.4 & \cellcolor{red!00}41.5 & \cellcolor{red!00}45.4 & \cellcolor{red!00}41.5 
    & \cellcolor{red!00}66.2 & \cellcolor{red!00}63.8 & \cellcolor{red!00}65.8 & \cellcolor{red!00}63.4 
    & \cellcolor{red!00}33.0 & \cellcolor{red!00}28.1 & \cellcolor{red!00}33.0 & \cellcolor{red!00}28.1 
    \\
    Query\&Extract
    & \cellcolor{red!00}-- & \cellcolor{red!00}-- & \cellcolor{red!00}-- & \cellcolor{red!00}-- 
    & \cellcolor{red!00}-- & \cellcolor{red!00}-- & \cellcolor{red!00}-- & \cellcolor{red!00}-- 
    & \cellcolor{red!00}52.2 & \cellcolor{red!00}50.3 & \cellcolor{red!00}51.8 & \cellcolor{red!00}50.0 
    & \cellcolor{red!00}-- & \cellcolor{red!00}-- & \cellcolor{red!00}-- & \cellcolor{red!00}-- 
    \\
    TagPrime-C
    & \cellcolor{red!24}70.4 & \cellcolor{red!24}65.7 & \cellcolor{red!24}68.6 & \cellcolor{red!16}64.0 
    & \cellcolor{red!24}54.4 & \cellcolor{red!16}48.3 & \cellcolor{red!24}54.4 & \cellcolor{red!16}48.3 
    & \cellcolor{red!24}83.0 & \cellcolor{red!24}79.2 & \cellcolor{red!24}82.7 & \cellcolor{red!24}79.0 
    & \cellcolor{red!16}55.3 & \cellcolor{red!16}54.4 & \cellcolor{red!16}55.3 & \cellcolor{red!16}54.4 
    \\
    TagPrime-CR
    & \cellcolor{red!24}70.3 & \cellcolor{red!24}67.2 & \cellcolor{red!16}68.4 & \cellcolor{red!24}65.5 
    & \cellcolor{red!16}54.1 & \cellcolor{red!24}49.7 & \cellcolor{red!16}54.1 & \cellcolor{red!24}49.7 
    & \cellcolor{red!24}82.8 & \cellcolor{red!24}80.4 & \cellcolor{red!24}82.5 & \cellcolor{red!24}80.1 
    & \cellcolor{red!24}55.5 & \cellcolor{red!24}54.7 & \cellcolor{red!24}55.5 & \cellcolor{red!24}54.7 
    \\
    DEGREE
    & \cellcolor{red!08}60.4 & \cellcolor{red!08}57.3 & \cellcolor{red!08}56.8 & \cellcolor{red!08}53.9 
    & \cellcolor{red!08}50.5 & \cellcolor{red!08}45.5 & \cellcolor{red!08}50.5 & \cellcolor{red!08}45.5 
    & \cellcolor{red!00}67.2 & \cellcolor{red!00}64.1 & \cellcolor{red!00}67.0 & \cellcolor{red!00}63.9 
    & \cellcolor{red!08}52.5 & \cellcolor{red!08}51.5 & \cellcolor{red!08}52.5 & \cellcolor{red!08}51.5 
    \\
    BART-Gen
    & \cellcolor{red!16}68.5 & \cellcolor{red!16}64.2 & \cellcolor{red!16}68.1 & \cellcolor{red!16}63.9 
    & \cellcolor{red!00}50.4 & \cellcolor{red!00}45.4 & \cellcolor{red!00}50.4 & \cellcolor{red!00}45.4 
    & \cellcolor{red!00}67.3 & \cellcolor{red!00}64.4 & \cellcolor{red!00}67.2 & \cellcolor{red!00}64.3 
    & \cellcolor{red!00}51.3 & \cellcolor{red!00}49.8 & \cellcolor{red!00}51.3 & \cellcolor{red!00}49.8 
    \\
    X-Gear
    & \cellcolor{red!00}58.7 & \cellcolor{red!00}55.6 & \cellcolor{red!00}55.4 & \cellcolor{red!00}52.4 
    & \cellcolor{red!16}52.1 & \cellcolor{red!08}46.2 & \cellcolor{red!16}52.1 & \cellcolor{red!08}46.2 
    & \cellcolor{red!16}78.9 & \cellcolor{red!16}75.1 & \cellcolor{red!16}78.7 & \cellcolor{red!16}74.9 
    & \cellcolor{red!08}51.5 & \cellcolor{red!08}50.4 & \cellcolor{red!08}51.5 & \cellcolor{red!08}50.4 
    \\
    PAIE
    & \cellcolor{red!16}69.8 & \cellcolor{red!16}65.5 & \cellcolor{red!24}69.5 & \cellcolor{red!24}65.2 
    & \cellcolor{red!24}55.2 & \cellcolor{red!24}50.5 & \cellcolor{red!24}55.2 & \cellcolor{red!24}50.5 
    & \cellcolor{red!16}73.5 & \cellcolor{red!16}70.4 & \cellcolor{red!16}73.4 & \cellcolor{red!16}70.3 
    & \cellcolor{red!00}48.8 & \cellcolor{red!00}47.9 & \cellcolor{red!00}48.8 & \cellcolor{red!00}47.9 
    \\
    Ampere
    & \cellcolor{red!08}59.9 & \cellcolor{red!08}56.7 & \cellcolor{red!08}56.2 & \cellcolor{red!08}53.3 
    & \cellcolor{red!08}52.0 & \cellcolor{red!16}46.8 & \cellcolor{red!08}52.0 & \cellcolor{red!16}46.8 
    & \cellcolor{red!08}67.8 & \cellcolor{red!08}65.0 & \cellcolor{red!08}67.6 & \cellcolor{red!08}64.8 
    & \cellcolor{red!00}-- & \cellcolor{red!00}-- & \cellcolor{red!00}-- & \cellcolor{red!00}-- 
    \\

    \bottomrule

\end{tabular}}
\caption{Reevaluation results for event argument extraction (EAE). All the numbers are the average score of 5 data splits. Darker cells imply higher scores. We use ``--'' to denote the cases that models are not runnable.}
\label{tab:all_eae}
\end{table*}

%% file: 93-prompt.tex
\section{Prompts for LLMs}
\label{app:prompts}

Table~\ref{tab:prompts} illustrates the prompts we use for testing the ability of LLMs in event detection and event argument extraction.

\begin{table*}[t!]
\centering
\setlength{\tabcolsep}{5pt}
\resizebox{.99\textwidth}{!}{
\begin{tabular}{cl}
    \toprule
    \multicolumn{2}{l}{\textbf{Prompt Used for Event Detection}}\\
    \midrule
    Instruction & \makecell[l]{
    You are an event extractor designed to check for the presence of a specific event in a sentence and to \\
    locate the corresponding event trigger.\\ 
    Task Description: Identify all triggers related to the event of interest in the sentence. A trigger is \\
    the key word in the sentence that most explicitly conveys the occurrence of the event. If yes, please \\
    answer `Yes, the event trigger is [trigger] in the text.'; otherwise, answer `No.' \\
    The event of interest is Business.Collaboration. This event is related to business collaboration.}\\
    \\[-0.8em]
    Example 1 & \makecell[l]{
    Examples 1\\
    Text: It is a way of coordinating different ideas from numerous people to generate a wide variety of \\ knowledge.\\
    Answer: Yes, the event trigger is \emph{coordinating} in the text.}\\
    \\[-0.8em]
    Example 2 & \makecell[l]{
    Examples 2\\
    Text: What's going on is that union members became outraged after learning about the airline's \\ 
    executive compensation plan where we would have paid huge bonuses even in bankruptcy\\
    Answer: No.}\\
    \\[-0.8em]
    ... & ...\\
    \\[-0.8em]
    Query & \makecell[l]{
    Question\\
    Text: Social networks permeate business culture where collaborative uses include file sharing and \\ knowledge transfer.\\
    Answer: }\\
    \\[-0.8em]
    Output & Yes, the event trigger is \emph{sharing} in the text.
    \\[-0.8em]
    \\
    \midrule
    \multicolumn{2}{l}{\textbf{Prompt Used for Event Argument Extraction}}\\
    \midrule
    Instruction & \makecell[l]{
    You are an argument extractor designed to check for the presence of arguments regarding specific \\
    roles for an event in a sentence. \\
    Task Description: Identify all arguments related to the role \emph{Agent}, \emph{Person}, \emph{Place} 
    in the sentence. \\
    These arguments should have the semantic role corresponding to the given event trigger by the word \\
    span between [t] and [/t]. Follow the the format of below examples. Your answer should only\\
    contain the answer string and nothing else.\\
    The event of interest is Justice:Arrest-Jail. The event is related to a person getting arrested or a \\ person being sent to jail. Roles of interest: \emph{Agent}, \emph{Person}, \emph{Place}}\\
    \\[-0.8em]
    Example 1 & \makecell[l]{
    Examples 1\\
    Text: Currently in California , 7000 people [t] serving [/t] 25 to year life sentences under the three\\
    strikes law.\\
    Agent:\\
    Person: people\\
    Place: California}\\
    \\[-0.8em]
    Example 2 & \makecell[l]{
    Examples 2\\
    Text: We've been playing warnings to people to stay in their houses , and we've only [t] lifted [/t] \\
    those people we've got very good intelligence on.\\
    Agent: we\\
    Person: people\\
    Place:}\\
    \\[-0.8em]
    ... & ... \\
    \\[-0.8em]
    Query & \makecell[l]{
    Question\\
    Text: A pizza delivery helped police [t] nab [/t] the suspect in the kidnapping of a 9-year-old \\
    California girl.}\\
    \\[-0.8em]
    Output & \makecell[l]{
    Agent: police\\
    Person: suspect\\
    Place:}\\
    \\[-0.8em]
    \bottomrule

\end{tabular}}
\caption{Prompts use for testing the ability of LLMs in event extraction.}
\label{tab:prompts}
\end{table*}